%% file: geotrans.tex
\documentclass[10pt,twocolumn,letterpaper]{article}

\usepackage{cvpr}              

\usepackage{graphicx}
\usepackage{amsmath}
\usepackage{amssymb}
\usepackage{booktabs}

\usepackage{multirow}
\usepackage{xcolor}
\usepackage{array}
\usepackage{multirow}
\usepackage{mathtools}
\usepackage{overpic}
\usepackage{stmaryrd}

\usepackage[pagebackref,breaklinks,colorlinks]{hyperref}

\usepackage[capitalize]{cleveref}
\crefname{section}{Sec.}{Secs.}
\Crefname{section}{Section}{Sections}
\crefname{table}{Tab.}{Tabs.}
\Crefname{table}{Table}{Tables}
\crefname{figure}{Fig.}{Figs.}
\Crefname{figure}{Figure}{Figures}
\crefname{equation}{Eq.}{Eqs.}
\Crefname{equation}{Equation}{Equations}
\crefname{appendix}{Appx.}{Appxs.}
\Crefname{Appendix}{Appendix}{Appendices}

\def\cvprPaperID{4196} 
\def\confName{CVPR}
\def\confYear{2022}

\input{hyphen}
\begin{document}

\title{Geometric Transformer for Fast and Robust Point Cloud Registration}

\author{
Zheng Qin$^1$\hspace{8pt}Hao Yu$^2$\hspace{8pt}Changjian Wang$^1$\hspace{8pt}Yulan Guo$^{1,3}$\hspace{8pt}Yuxing Peng$^1$\hspace{8pt}Kai Xu$^1$\thanks{Corresponding author: kevin.kai.xu@gmail.com.}\\
\hspace{-8pt}$^1$National University of Defense Technology\hspace{5pt}$^2$Technical University of Munich\hspace{5pt}$^3$Sun Yat-sen University
}
\maketitle

\input{abstract}


\input{intro}
\input{related}
\input{method}
\input{results}
\input{conclusion}

\clearpage
\input{appendix}

{\small
\bibliographystyle{ieee_fullname}
\bibliography{geotrans}
}


\end{document}

%% file: hyphen.tex
\usepackage{amsmath}

\DeclareMathOperator*{\argmin}{\arg\!\min}

\def\tight#1{\hspace{1pt}{#1}{\hspace{1pt}}}
\def\medium#1{\hspace{2pt}{#1}{\hspace{2pt}}}

\newcolumntype{P}[1]{>{\centering\arraybackslash}p{#1}}
\newlength{\wdth}

\newcommand{\ptitle}[1]{\noindent\textbf{#1}\hspace{5pt}}

%% file: abstract.tex

\begin{abstract}
We study the problem of extracting accurate correspondences for point cloud registration.
Recent keypoint-free methods bypass the detection of repeatable keypoints which is difficult in low-overlap scenarios, showing great potential in registration. They seek correspondences over downsampled superpoints, which are then propagated to dense points.
Superpoints are matched based on whether their neighboring patches overlap.
Such sparse and loose matching requires contextual features capturing the geometric structure of the point clouds. We propose Geometric Transformer to learn geometric feature for robust superpoint matching. It encodes pair-wise distances and triplet-wise angles, making it robust in low-overlap cases and invariant to rigid transformation.
The simplistic design attains surprisingly high matching accuracy such that no RANSAC is required in the estimation of alignment transformation, leading to $100$ times acceleration. Our method improves the inlier ratio by $17{\sim}30$ percentage points and the registration recall by over $7$ points on the challenging 3DLoMatch benchmark.
Our code and models are available at \url{https://github.com/qinzheng93/GeoTransformer}.
\end{abstract}
\vspace{-10pt}

%% file: intro.tex

\section{Introduction}
Point cloud registration is a fundamental task in graphics, vision and robotics.
Given two partially overlapping 3D point clouds, the goal is to estimate a rigid transformation that aligns them. The problem has gained renewed interest recently thanks to the fast growing of 3D point representation learning and differentiable optimization.


The recent advances have been dominated by learning-based, correspondence-based methods~\cite{deng2018ppfnet,gojcic2019perfect,choy2019fully,bai2020d3feat,huang2021predator,yu2021cofinet}. A neural network is trained to extract point correspondences between two input point clouds, based on which an alignment transformation is calculated with a robust estimator, \eg, RANSAC. 
Most correspondence-based methods rely on keypoint detection~\cite{choy2019fully,bai2020d3feat,ao2021spinnet,huang2021predator}.
However, it is challenging to detect repeatable keypoints across two point clouds, especially when they have small overlapping area. This usually results in low inlier ratio in the putative correspondences.

Inspired by the recent advances in image matching~\cite{rocco2018neighbourhood,zhou2021patch2pix,sun2021loftr},
keypoint-free methods~\cite{yu2021cofinet} downsample the input point clouds into superpoints and then match them through examining whether their local neighborhood (patch) overlaps. Such superpoint (patch) matching is then propagated to individual points, yielding dense point correspondences. Consequently, the accuracy of dense point correspondences highly depends on that of superpoint matches.

\input{figures/teaser}

Superpoint matching is sparse and loose. The upside is that it reduces strict point matching into loose patch overlapping, thus relaxing the repeatability requirement. Meanwhile, patch overlapping is a more reliable and informative constraint than distance-based point matching for learning correspondence; consider that two spatially close points could be geodesically distant. On the other hand, superpoint matching calls for features capturing more global context.

To this end, Transformer~\cite{vaswani2017attention} has been adopted~\cite{wang2019deep,yu2021cofinet} to encode contextual information in point cloud registration. However, vanilla transformer overlooks the geometric structure of the point clouds, which makes the learned features geometrically less discriminative and induces numerous outlier matches (\cref{fig:teaser}(top)). Although one can inject positional embeddings~\cite{zhao2021point,yang2019modeling}, the coordinate-based encoding is transformation-variant, which is problematic when registering point clouds given in arbitrary poses.
We advocate that a point transformer for registration task should be learned with the \emph{geometric structure} of the point clouds so as to extract transformation-invariant geometric features. We propose \emph{Geometric Transformer}, or \emph{GeoTransformer} for short, for 3D point clouds which encodes only distances of point pairs and angles in point triplets.

Given a superpoint, we learn a non-local representation through geometrically ``pinpointing'' it w.r.t. all other superpoints based on pair-wise distances and triplet-wise angles. Self-attention mechanism is utilized to weigh the importance of those anchoring superpoints.
Since distances and angles are invariant to rigid transformation, GeoTransformer learns geometric structure of point clouds efficiently, leading to highly robust superpoint matching even in low-overlap scenarios. \cref{fig:teaser}(left) demonstrates that GeoTransformer significantly improves the inlier ratio of superpoint (patch) correspondences. For better convergence, we devise an overlap-aware circle loss to make GeoTransformer focus on superpoint pairs with higher patch overlap.

Benefitting from the high-quality superpoint matches, our method attains high-inlier-ratio dense point correspondences (\cref{fig:teaser}(right)) using an optimal transport layer~\cite{sarlin2020superglue}, as well as highly robust and accurate registration without relying on RANSAC. Therefore, the registration part of our method runs extremely fast, \eg, $0.01$s for two point clouds with $5$K correspondences, $100$ times faster than RANSAC.
Extensive experiments on both indoor and outdoor benchmarks~\cite{zeng20173dmatch,geiger2012we} demonstrate the efficacy of GeoTransformer.
Our method improves the inlier ratio by $17{\sim}30$ percentage points and the registration recall by over $7$ points on the 3DLoMatch benchmark~\cite{huang2021predator}.
Our main contributions are:
\begin{itemize}
  \vspace{-8pt}
  \item A fast and accurate point cloud registration method which is both keypoint-free and RANSAC-free.
  \vspace{-8pt}
  \item A geometric transformer which learns transformation-invariant geometric representation of point clouds for robust superpoint matching.
  \vspace{-8pt}
  \item An overlap-aware circle loss which reweights the loss of each superpoint match according to the patch overlap ratio for better convergence.
\end{itemize}

%% file: figures/teaser.tex

\begin{figure}[t]
  \begin{overpic}[width=1.0\linewidth]{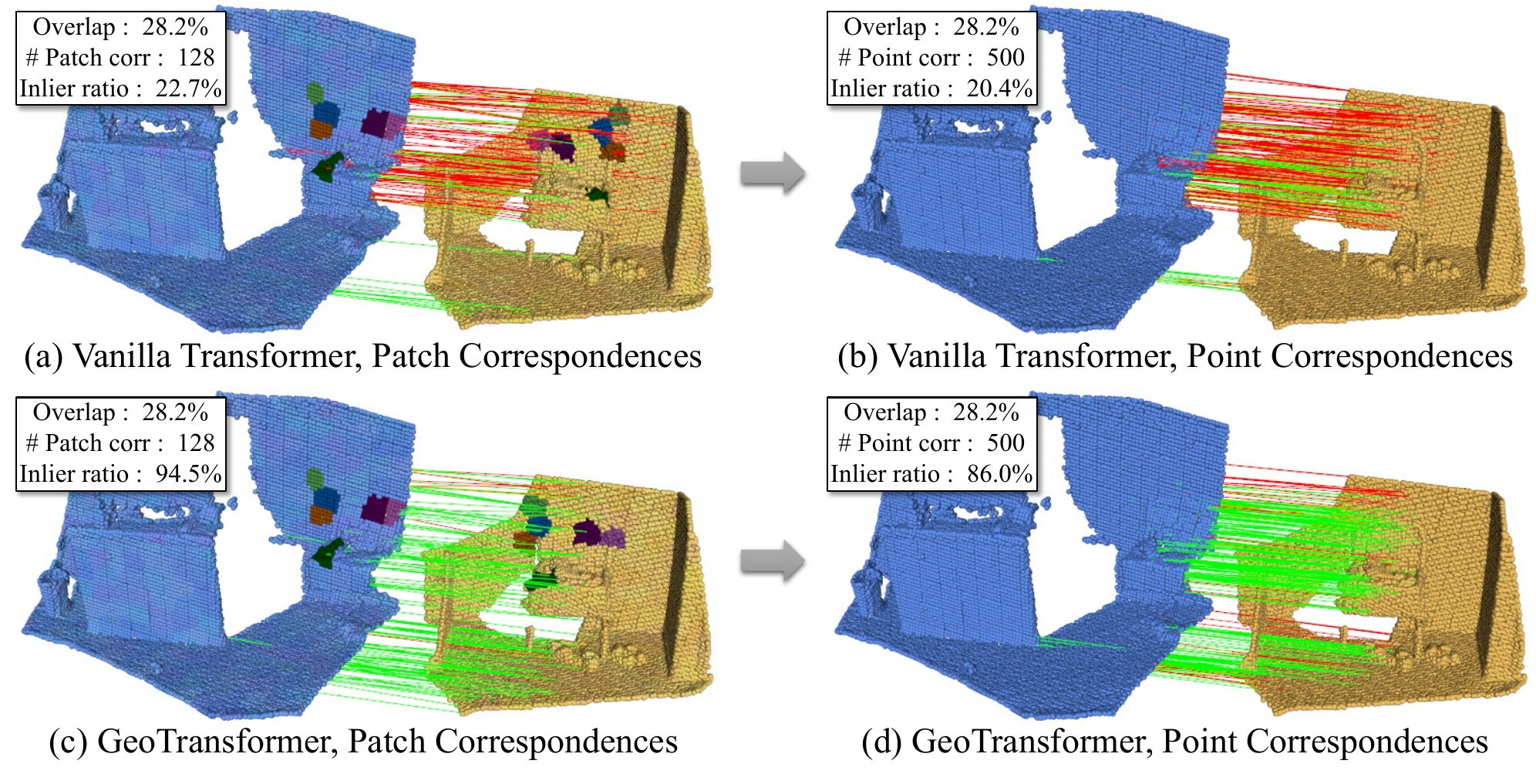}
  \end{overpic}
  \vspace{-20pt}
  \caption{Given two low-overlap point clouds, GeoTransformer improves inlier ratio over vanilla transformer significantly, both for superpoint (patch) level (left) and for dense point level (right). A few representative patch correspondences are visualized with distinct colors. Notice how GeoTransformer preserves the spatial consistency of the matching patches across two point clouds. It corrects the wrongly matched patches around the symmetric corners of the chair back (see the yellow point cloud).}
  \label{fig:teaser}
  \vspace{-10pt}
\end{figure}

%% file: related.tex

\section{Related Work}
\label{sec:related}

\ptitle{Correspondence-based Methods.}
Our work follows the line of the correspondence-based methods~\cite{deng2018ppfnet,deng2018ppf,gojcic2019perfect,choy2019fully}.
They first extract correspondences between two point clouds and then recover the transformation with robust pose estimators, \eg, RANSAC.
Thanks to the robust estimators, they achieve state-of-the-art performance in indoor and outdoor scene registration.
These methods can be further categorized into two classes according to how they extract correspondences.
The first class aims to detect more repeatable keypoints~\cite{bai2020d3feat,huang2021predator} and learn more powerful descriptors for the keypoints~\cite{choy2019fully,ao2021spinnet,wang2021you}.
While the second class~\cite{yu2021cofinet} retrieves correspondences without keypoint detection by considering all possible matches.
Our method follows the detection-free methods and improves the accuracy of correspondences by leveraging the geometric information.

\ptitle{Direct Registration Methods.}
Recently, direct registration methods have emerged. They estimate the transformation with a neural network in an end-to-end manner.
These methods can be further classified into two classes.
The first class~\cite{wang2019deep,wang2019prnet,yew2020rpm,fu2021robust} follows the idea of ICP~\cite{besl1992method}, which iteratively establishes soft correspondences and computes the transformation with differentiable weighted SVD.
The second class~\cite{aoki2019pointnetlk,huang2020feature,xu2021omnet} first extracts a global feature vector for each point cloud and regresses the transformation with the global feature vectors.
Although direct registration methods have achieved promising results on single synthetic shapes, they could fail in large-scale scenes as stated in~\cite{huang2021predator}.

\ptitle{Deep Robust Estimators.}
As traiditional robust estimators such as RANSAC suffer from slow convergence and instability in case of high outlier ratio, deep robust estimators~\cite{pais20203dregnet,choy2020deep,bai2021pointdsc} have been proposed as the alternatives for them.
They usually contain a classification network to reject outliers and an estimation network to compute the transformation.
Compared with traditional robust estimators, they achieve improvements in both accuracy and speed.
However, they require training a specific network.
In comparison, our method achieves fast and accurate registration with a parameter-free local-to-global registration scheme.

%% file: method.tex

\section{Method}
\label{sec:geotrans}

\input{figures/pipeline}

Given two point clouds $\mathcal{P} = \{\textbf{p}_i \in \mathbb{R}^3 \mid i = 1, ..., N\}$ and $\mathcal{Q} = \{\textbf{q}_i \in \mathbb{R}^3 \mid i = 1, ..., M\}$, our goal is to estimate a rigid transformation $\textbf{T} = \{\textbf{R}, \textbf{t}\}$ which aligns the two point clouds, with a 3D rotation $\textbf{R} \in SO(3)$ and a 3D translation $\textbf{t} \in \mathbb{R}^3$.
The transformation can be solved by:
\begin{equation}
\min_{\textbf{R}, \textbf{t}} \sum\nolimits_{(\textbf{p}^{*}_{x_i}, \textbf{q}^{*}_{y_i}) \in \mathcal{C}^{*}} \lVert \textbf{R} \cdot \textbf{p}^{*}_{x_i} + \textbf{t} - \textbf{q}^{*}_{y_i} \rVert^2_2.
\end{equation}
Here $\mathcal{C}^{*}$ is the set of ground-truth correspondences between $\mathcal{P}$ and $\mathcal{Q}$.
Since $\mathcal{C}^{*}$ is unknown in reality, we need to first establish point correspondences between two point clouds and then estimate the alignment transformation.

Our method adopts the hierarchical correspondence paradigm which finds correspondences in a coarse-to-fine manner.
We adopt KPConv-FPN to simultaneously downsample the input point clouds and extract point-wise features (\cref{sec:model-backbone}). The first and the last (coarsest) level downsampled points correspond to the dense points and the superpoints to be matched.
A \emph{Superpoint Matching Module} is used to extract superpoint correspondences whose neighboring local patches overlap with each other (\cref{sec:model-pam}).
Based on that, a \emph{Point Matching Module} then refines the superpoint correspondences to dense points (\cref{sec:model-pom}). At last, the alignment transformation is recovered from the dense correspondences without relying on RANSAC (\cref{sec:model-estimation}).
The pipeline is illustrated in \cref{fig:overview}.

\subsection{Superpoint Sampling and Feature Extraction}
\label{sec:model-backbone}
We utilize the KPConv-FPN backbone~\cite{thomas2019kpconv,lin2017feature} to extract multi-level features for the point clouds. A byproduct of the point feature learning is point downsampling. We work on downsampled points since point cloud registration can actually be pinned down by the correspondences of a much coarser subset of points. The original point clouds are usually too dense so that point-wise correspondences are redundant
and sometimes too clustered to be useful.

The points correspond to the coarsest resolution, denoted by $\hat{\mathcal{P}}$ and $\hat{\mathcal{Q}}$, are treated as \emph{superpoints} to be matched. The associated learned features are denoted as $\hat{\textbf{F}}{}^{\mathcal{P}} \medium{\in} \mathbb{R}^{\lvert \hat{\mathcal{P}} \rvert \times \hat{d}}$ and $\hat{\textbf{F}}{}^{\mathcal{Q}} \medium{\in} \mathbb{R}^{\lvert \hat{\mathcal{Q}} \rvert \times \hat{d}}$.
The dense point correspondences are computed at $1/2$ of the original resolution, \ie, the first level downsampled points denoted by $\tilde{\mathcal{P}}$ and $\tilde{\mathcal{Q}}$. Their learned features are represented by $\tilde{\textbf{F}}{}^{\mathcal{P}} \medium{\in} \mathbb{R}^{\lvert \tilde{\mathcal{P}} \rvert \times \tilde{d}}$ and $\tilde{\textbf{F}}{}^{\mathcal{Q}} \medium{\in} \mathbb{R}^{\lvert \tilde{\mathcal{Q}} \rvert \times \tilde{d}}$.

For each superpoint, we construct a local \emph{patch} of points around it using the point-to-node grouping strategy \cite{li2018so,yu2021cofinet}.
In particular, each point in $\tilde{\mathcal{P}}$ and its features from $\tilde{\textbf{F}}{}^{\mathcal{P}}$ are assigned to its nearest superpoint in the geometric space:
\vspace{-2pt}
\begin{equation}
\mathcal{G}^{\mathcal{P}}_i = \{\tilde{\textbf{p}} \in \tilde{\mathcal{P}} \mid i = \argmin\nolimits_j(\lVert \tilde{\textbf{p}} - \hat{\textbf{p}}_j \rVert_2), \hat{\textbf{p}}_j \in \hat{\mathcal{P}}\}.
\vspace{-2pt}
\end{equation}
This essentially leads to a Voronoi decomposition of the input point cloud seeded by superpoints.
The feature matrix associated with the points in $\mathcal{G}^{\mathcal{P}}_i$ is denoted as $\textbf{F}^{\mathcal{P}}_i \subset \tilde{\textbf{F}}{}^{\mathcal{P}}$.
The superpoints with an empty patch are removed.
The patches $\{\mathcal{G}^{\mathcal{Q}}_i\}$ and the feature matrices $\{\textbf{F}^{\mathcal{Q}}_i\}$ for $\mathcal{Q}$ are computed and denoted in a similar way.

\subsection{Superpoint Matching Module}
\label{sec:model-pam}

\input{figures/geotr}

\ptitle{Geometric Transformer.}
Global context has proven critical in many computer vision tasks~\cite{dosovitskiy2020image,sun2021loftr,yu2021cofinet}.
For this reason, transformer has been adopted to leverage global contextual information for point cloud registration.
However, existing methods~\cite{wang2019deep,huang2021predator,yu2021cofinet} usually feed transformer with only high-level point cloud features and does not explicitly encode the geometric structure.
This makes the learned features geometrically less discriminative, which causes severe matching ambiguity and numerous outlier matches, especially in low-overlap cases.
A straightforward recipe is to explicitly inject positional embeddings~\cite{yang2019modeling,zhao2021point} of 3D point coordinates.
However, the resultant coordinate-based transformers are naturally \emph{transformation-variant}, while registration requires \emph{transformation invariance} since the input point clouds can be in arbitrary poses.

To this end, we propose \emph{Geometric Transformer} which not only encodes high-level point features but also explicitly captures intra-point-cloud geometric structures and inter-point-cloud geometric consistency.
GeoTransformer is composed of a \emph{geometric self-attention} module for learning intra-point-cloud features and a \emph{feature-based cross-attention} module for modeling inter-point-cloud consistency. The two modules are interleaved for $N_t$ times to extract hybrid features $\hat{\textbf{H}}{}^{\mathcal{P}}$ and $\hat{\textbf{H}}{}^{\mathcal{Q}}$ for reliable superpoint matching (see \cref{fig:overview} (bottom left)).

\ptitle{Geometric self-attention.}
We design a \emph{geometric self-attention} to learn the global correlations in both feature and geometric spaces among the superpoints for each point cloud.
In the following, we describe the computation for $\hat{\mathcal{P}}$ and the same goes for $\hat{\mathcal{Q}}$. Given the input feature matrix $\textbf{X} \medium{\in} \mathbb{R}^{\lvert \hat{\mathcal{P}} \vert \times d_t}$, the output feature matrix $\textbf{Z} \medium{\in} \mathbb{R}^{\lvert \hat{\mathcal{P}} \vert \times d_t}$ is the weighted sum of all projected input features:
\vspace{-5pt}
\begin{equation}
\textbf{z}_i = \sum_{j=1}^{\lvert \hat{\mathcal{P}} \vert} a_{i, j} (\textbf{x}_j\textbf{W}^V),
\vspace{-5pt}
\end{equation}
where the weight coefficient $a_{i, j}$ is computed by a row-wise softmax on the attention score $e_{i, j}$, and $e_{i, j}$ is computed as:
\vspace{-5pt}
\begin{equation}
e_{i, j} = \frac{(\textbf{x}_i\textbf{W}^Q)(\textbf{x}_j\textbf{W}^K + \textbf{r}_{i, j}\textbf{W}^R)^T}{\sqrt{d_{t}}}.
\vspace{-5pt}
\end{equation}
Here, $\textbf{r}_{i, j} \medium{\in} \mathbb{R}^{d_t}$ is a \emph{geometric structure embedding} to be described in the next. $\textbf{W}^Q, \textbf{W}^K, \textbf{W}^V, \textbf{W}^R \in \mathbb{R}^{d_t \times d_t}$ are the respective projection matrices for queries, keys, values and geometric structure embeddings.
\cref{fig:geotr} shows the structure and the computation of geometric self-attention.

We design a novel \emph{geometric structure embedding} to encode the transformation-invariant geometric structure of the superpoints.
The core idea is to leverage the distances and angles computed with the superpoints which are consistent across different point clouds of the same scene.
Given two superpoints $\hat{\textbf{p}}_i, \hat{\textbf{p}}_j \tight{\in} \hat{\mathcal{P}}$, their geometric structure embedding consists of a \emph{pair-wise distance embedding} and a \emph{triplet-wise angular embedding}, which will be described below.

(1)
\emph{Pair-wise Distance Embedding}.
Given the distance $\rho_{i, j} \hspace{1pt} {=} \hspace{1pt} \lVert \hat{\textbf{p}}_i - \hat{\textbf{p}}_j \rVert_2$ between $\hat{\textbf{p}}_i$ and $\hat{\textbf{p}}_j$, the distance embedding $\textbf{r}^D_{i, j}$ between them is computed by applying a sinusoidal function \cite{vaswani2017attention} on $\rho_{i, j} / \sigma_d$.
Here, $\sigma_d$ is a hyper-parameter used to tune the sensitivity on distance variations.
Please refer to the Appx.~\textcolor{red}{A.1} for detailed computation.

(2) \emph{Triplet-wise Angular Embedding}.
We compute angular embedding with triplets of superpoints.
We first select the $k$ nearest neighbors $\mathcal{K}_i$ of $\hat{\textbf{p}}_i$.
For each $\hat{\textbf{p}}_x \tight{\in} \mathcal{K}_i$, we compute the angle $\alpha^x_{i,j} \tight{=} \angle(\Delta_{x, i}, \Delta_{j, i})$, where $\Delta_{i, j} \tight{:=} \hat{\textbf{p}}_i \hspace{1pt} {-} \hspace{1pt} \hat{\textbf{p}}_j$.
The triplet-wise angular embedding $\textbf{r}^A_{i, j, x}$ is then computed with a sinusoidal function on $\alpha^x_{i,j} / \sigma_a$, with $\sigma_a$ controlling the sensitivity on angular variations.

Finally, the geometric structure embedding $\textbf{r}_{i, j}$ is computed by aggregating the pair-wise distance embedding and the triplet-wise angular embedding:
\vspace{-5pt}
\begin{equation}
\textbf{r}_{i, j} = \textbf{r}^D_{i, j}\textbf{W}^D + {\max}_x\left\{\textbf{r}^A_{i, j, x}\textbf{W}^A\right\},
\vspace{-5pt}
\label{eq:gse}
\end{equation}
where $\textbf{W}^D, \textbf{W}^A \in \mathbb{R}^{d_t \times d_t}$ are the respective projection matrices for the two types of embeddings. We use max pooling here to improve the robustness to the varying nearest neighbors of a superpoint due to self-occlusion.
\cref{fig:rge} illustrates the computation of geometric structure embedding.


\input{figures/rge}

\ptitle{Feature-based cross-attention.}
Cross-attention is a typical module for point cloud registration task~\cite{huang2021predator,wang2019deep,yu2021cofinet}, used to perform feature exchange between two input point clouds.
Given the self-attention feature matrices $\textbf{X}^{\mathcal{P}}$, $\textbf{X}^{\mathcal{Q}}$ for $\hat{\mathcal{P}}$, $\hat{\mathcal{Q}}$ respectively, the cross-attention feature matrix $\textbf{Z}^{\mathcal{P}}$ of $\hat{\mathcal{P}}$ is computed with the features of $\hat{\mathcal{Q}}$:
\vspace{-2pt}
\begin{equation}
\textbf{z}^{\mathcal{P}}_i = \sum_{j=1}^{\lvert \hat{\mathcal{Q}} \rvert} a_{i, j} (\textbf{x}^{\mathcal{Q}}_j\textbf{W}^V).
\vspace{-2pt}
\end{equation}
Similarly, $a_{i, j}$ is computed by a row-wise softmax on the cross-attention score $e_{i, j}$, and $e_{i, j}$ is computed as the feature correlation between the $\textbf{X}^{\mathcal{P}}$ and $\textbf{X}^{\mathcal{Q}}$:
\vspace{-2pt}
\begin{equation}
e_{i, j} = \frac{(\textbf{x}^{\mathcal{P}}_i\textbf{W}^Q)(\textbf{x}^{\mathcal{Q}}_j\textbf{W}^K)^T}{\sqrt{d_{t}}}.
\vspace{-2pt}
\end{equation}
The cross-attention features for $\mathcal{Q}$ are computed in the same way.
While the geometric self-attention module encodes the transformation-invariant geometric structure for each individual point cloud, the feature-based cross-attention module can model the geometric consistency across the two point clouds.
The resultant hybrid features are both invariant to transformation and robust for reasoning correspondence.

\ptitle{Superpoint matching.}
To find the superpoint correspondences, we propose a matching scheme based on global feature correlation.
We first normalize $\hat{\textbf{H}}{}^{\mathcal{P}}$ and $\hat{\textbf{H}}{}^{\mathcal{Q}}$ onto a unit hypersphere and compute a Gaussian correlation matrix $\textbf{S} \tight{\in} \mathbb{R}^{\lvert \hat{\mathcal{P}} \rvert \times \lvert \hat{\mathcal{Q}} \rvert}$ with $s_{i, j} \tight{=} \exp(-\lVert \hat{\textbf{h}}{}^{\mathcal{P}}_i \tight{-} \hat{\textbf{h}}{}^{\mathcal{Q}}_j\rVert_2^2)$.
In practice, some patches of a point cloud are less geometrically discriminative and have numerous similar patches in the other point cloud. Besides our powerful hybrid features, we also perform a dual-normalization operation \cite{rocco2018neighbourhood,sun2021loftr} on $\textbf{S}$ to further suppress ambiguous matches, leading to $\bar{\textbf{S}}$ with
\vspace{-2pt}
\begin{equation}
\bar{s}_{i, j} = \frac{s_{i, j}}{\sum_{k=1}^{\lvert \hat{\mathcal{Q}} \rvert} s_{i, k}} \cdot \frac{s_{i, j}}{\sum_{k=1}^{\lvert \hat{\mathcal{P}} \rvert} s_{k, j}}.
\vspace{-2pt}
\end{equation}
We found that this suppression can effectively eliminate wrong matches.
Finally, we select the largest $N_{c}$ entries in $\bar{\textbf{S}}$ as the \emph{superpoint correspondences}:
\begin{equation}
\hat{\mathcal{C}} = \{ (\hat{\textbf{p}}_{x_i}, \hat{\textbf{q}}_{y_i}) \mid (x_i, y_i) \in \mathrm{topk}_{x, y}(\bar{s}_{x, y}) \}.
\end{equation}
Due to the powerful geometric structure encoding of GeoTransformer, our method is able to achieve accurate registration in low-overlap cases and with less point correspondences, and most notably, in a RANSAC-free manner.

\subsection{Point Matching Module}
\label{sec:model-pom}

Having obtained the superpoint correspondences, we extract point correspondences using a simple yet effective \emph{Point Matching Module}.
At point level, we use only local point features learned by the backbone.
The rationale is that point level matching is mainly determined by the vicinities of the two points being matched, once the global ambiguity has been resolved by superpoint matching.
This design choice improves the robustness.

For each superpoint correspondence $\hat{\mathcal{C}}_i = (\hat{\textbf{p}}_{x_i}, \hat{\textbf{q}}_{y_i})$, an optimal transport layer \cite{sarlin2020superglue} is used to extract the \emph{local dense point correspondences} between $\mathcal{G}^{\mathcal{P}}_{x_i}$ and $\mathcal{G}^{\mathcal{Q}}_{y_i}$.
Specifically, we first compute a cost matrix $\textbf{C}_i \in \mathbb{R}^{n_i \times m_i}$:
\vspace{-2pt}
\begin{equation}
\textbf{C}_i = \textbf{F}^{\mathcal{P}}_{x_i} (\textbf{F}^{\mathcal{Q}}_{y_i})^T / \sqrt{\tilde{d}},
\vspace{-2pt}
\end{equation}
where $n_i = \lvert \mathcal{G}^{\mathcal{P}}_{x_i} \rvert$, $m_i = \lvert \mathcal{G}^{\mathcal{Q}}_{y_i} \rvert$.
The cost matrix $\textbf{C}_i$ is then augmented into $\bar{\textbf{C}}_i$ by appending a new row and a new column as in \cite{sarlin2020superglue}, filled with a learnable dustbin parameter $\alpha$.
We then utilize the Sinkhorn algorithm \cite{sinkhorn1967concerning} on $\bar{\textbf{C}}_i$ to compute a soft assignment matrix $\bar{\textbf{Z}}_i$ which is then recovered to $\textbf{Z}_i$ by dropping the last row and the last column.
We use $\textbf{Z}_i$ as the confidence matrix of the candidate matches and extract point correspondences via mutual top-$k$ selection, where a point match is selected if it is among the $k$ largest entries of both the row and the column that it resides in:
\vspace{-2pt}
\begin{equation}
\mathcal{C}_i \tight{=} \{(\mathcal{G}^{\mathcal{P}}_{x_i}(x_j), \mathcal{G}^{\mathcal{Q}}_{y_i}(y_j)) \tight{\mid} (x_j, y_j) \tight{\in} \mathrm{mutual\_topk}_{x, y}(z^i_{x, y})\}.
\end{equation}
The point correspondences computed from each superpoint match are then collected together to form the final \emph{global dense point correspondences}: $\mathcal{C} = \bigcup_{i=1}^{N_c} \mathcal{C}_i$.

\subsection{RANSAC-free Local-to-Global Registration}
\label{sec:model-estimation}

Previous methods generally rely on robust pose estimators to estimate the transformation since the putative correspondences are often predominated by outliers.
Most robust estimators such as RANSAC suffer from slow convergence.
Given the high inlier ratio of GeoTransformer, we are able to achieve robust registration without relying on robust estimators, which also greatly reduces computation cost.

We design a \emph{local-to-global registration} (LGR) scheme.
As a hypothesize-and-verify approach, LGR is comprised of a local phase of transformation candidates generation and a global phase for transformation selection.
In the local phase, we solve for a transformation $\textbf{T}_i \tight{=} \{\textbf{R}_i, \textbf{t}_i\}$ for each superpoint match using its \emph{local point correspondences}:
\vspace{-2pt}
\begin{equation}
\textbf{R}_i, \textbf{t}_i = \min_{\textbf{R}, \textbf{t}} \sum\nolimits_{(\tilde{\textbf{p}}_{x_j}, \tilde{\textbf{q}}_{y_j}) \in \mathcal{C}_i} w^i_j \lVert \textbf{R} \cdot \tilde{\textbf{p}}_{x_j} \hspace{-3pt} + \textbf{t} - \tilde{\textbf{q}}_{y_j} \rVert_2^2.
\label{eq:weighted-svd}
\vspace{-2pt}
\end{equation}
This can be solved in closed form using weighted SVD~\cite{besl1992method}.
The corresponding confidence score for each correspondence in $\textbf{Z}_i$ is used as the weight $w^i_j$.
Benefitting from the high-quality correspondences, the transformations obtained in this phase are already very accurate.
In the global phase, we select the transformation which admits the most inlier matches over the entire \emph{global point correspondences}:
\vspace{-2pt}
\begin{equation}
\textbf{R}, \textbf{t} = \max_{\textbf{R}_i, \textbf{t}_i} \sum\nolimits_{(\tilde{\textbf{p}}_{x_j}, \tilde{\textbf{q}}_{y_j}) \in \mathcal{C}} \llbracket \lVert \textbf{R}_i \cdot \tilde{\textbf{p}}_{x_j} \hspace{-3pt} + \textbf{t}_i - \tilde{\textbf{q}}_{y_j} \rVert_2^2 < \tau_a \rrbracket,
\end{equation}
where $\llbracket \cdot \rrbracket$ is the Iverson bracket. $\tau_a$ is the acceptance radius.
We then iteratively re-estimate the transformation with the surviving inlier matches for $N_r$ times by solving \cref{eq:weighted-svd}.
As shown in \cref{sec:exp-indoor}, our approach achieves comparable registration accuracy with RANSAC but reduces the computation time by more than $100$ times.
Moreover, unlike deep robust estimators~\cite{choy2020deep,pais20203dregnet,bai2021pointdsc}, our method is parameter-free and no network training is needed.

\subsection{Loss Functions}
\label{sec:model-loss}

The loss function $\mathcal{L} = \mathcal{L}_{oc} + \mathcal{L}_{p}$ is composed of an \emph{overlap-aware circle loss} $\mathcal{L}_{oc}$ for superpoint matching and a \emph{point matching loss} $\mathcal{L}_{p}$ for point matching.

\ptitle{Overlap-aware circle loss.}
Existing methods~\cite{sun2021loftr,yu2021cofinet} usually formulate superpoint matching as a multi-label classification problem and adopt a cross-entropy loss with dual-softmax~\cite{sun2021loftr} or optimal transport~\cite{sarlin2020superglue,yu2021cofinet}.
Each superpoint is assigned (classified) to one or many of the other superpoints, where the ground truth is computed based on patch overlap and it is very likely that one patch could overlap with multiple patches.
By analyzing the gradients from the cross-entropy loss, we find that the positive classes with high confidence scores are suppressed by positive gradients in the multi-label classification\footnote{The detailed analysis is presented in Appx.~\textcolor{red}{C}.}.
This hinders the model from extracting reliable superpoint correspondences.

To address this issue, we opt to extract superpoint descriptors in a metric learning fashion.
A straightforward solution is to adopt a circle loss~\cite{sun2020circle} similar to~\cite{bai2020d3feat,huang2021predator}.
However, the circle loss overlooks the differences between the positive samples and weights them equally.
As a result, it struggles in matching patches with relatively low overlap.
For this reason, we design an \emph{overlap-aware circle loss} to focus the model on those matches with high overlap.
We select the patches in $\mathcal{P}$ which have at least one positive patch in $\mathcal{Q}$ to form a set of anchor patches, $\mathcal{A}$.
A pair of patches are positive if they share at least $10\%$ overlap, and negative if they do not overlap.
All other pairs are omitted.
For each anchor patch $\mathcal{G}^{\mathcal{P}}_i \in \mathcal{A}$, we denote the set of its positive patches in $\mathcal{Q}$ as $\varepsilon^i_p$, and the set of its negative patches as $\varepsilon^i_n$.
The overlap-aware circle loss on $\mathcal{P}$ is then defined as:
\begin{equation}
\label{eq:overlap-aware-circle-loss}
\mathcal{L}^{\mathcal{P}}_{oc} \tight{=} \frac{1}{\lvert\mathcal{A}\rvert} \sum_{\mathclap{\mathcal{G}^{\mathcal{P}}_i \in \mathcal{A}}} \log[1 + \sum_{\mathclap{\mathcal{G}^{\mathcal{Q}}_j \in \varepsilon^i_p}} e^{\lambda^j_i\beta^{i,j}_p (d^j_i - \Delta_p)} \cdot \sum_{\mathclap{\mathcal{G}^{\mathcal{Q}}_k \in \varepsilon^i_n}} e^{\beta^{i,k}_n (\Delta_n - d^k_i)}],
\end{equation}
where $d^j_i \hspace{1pt} {=} \hspace{1pt} \lVert \hat{\textbf{h}}{}^{\mathcal{P}}_i \hspace{1pt} {-} \hspace{1pt} \hat{\textbf{h}}{}^{\mathcal{Q}}_j \rVert_2$ is the distance in the feature space, $\lambda_i^j \tight{=} (o^j_i)^{\frac{1}{2}}$ and $o^j_i$ represents the overlap ratio between $\mathcal{G}^{\mathcal{P}}_i$ and $\mathcal{G}^{\mathcal{Q}}_j$.
The positive and negative weights are computed for each sample individually with $\beta^{i,j}_p \medium{=} \gamma(d^j_i \medium{-} \Delta_p)$ and $\beta^{i,k}_n \medium{=} \gamma(\Delta_n \medium{-} d^k_i)$.
The margin hyper-parameters are set to $\Delta_p \hspace{1pt} {=} \hspace{1pt} 0.1$ and $\Delta_n \hspace{1pt} {=} \hspace{1pt} 1.4$.
The overlap-aware circle loss reweights the loss values on $\varepsilon^i_p$ based on the overlap ratio so that the patch pairs with higher overlap are given more importance.
The same goes for the loss $\mathcal{L}^{\mathcal{Q}}_{oc}$ on $\mathcal{Q}$. And the overall loss is $\mathcal{L}_{oc} = (\mathcal{L}^{\mathcal{P}}_{oc} + \mathcal{L}^{\mathcal{Q}}_{oc}) / 2$.

\ptitle{Point matching loss.}
The ground-truth point correspondences are relatively sparse because they are available only for downsampled point clouds.
We simply use a negative log-likelihood loss~\cite{sarlin2020superglue} on the assignment matrix $\bar{\textbf{Z}}_i$ of each superpoint correspondence.
During training, we randomly sample $N_g$ ground-truth superpoint correspondences $\{\hat{\mathcal{C}}^{*}_i\}$ instead of using the predicted ones.
For each $\hat{\mathcal{C}}^{*}_i$, a set of ground-truth point correspondences $\mathcal{M}_i$ is extracted with a matching radius $\tau$. The sets of unmatched points in the two patches are denoted as $\mathcal{I}_i$ and $\mathcal{J}_i$.
The individual point matching loss for $\hat{\mathcal{C}}^{*}_i$ is computed as:
\begin{equation}
\mathcal{L}_{p, i} = -\sum_{\mathclap{{(x, y) \in \mathcal{M}_i}}} \log \bar{z}^i_{x, y} - \sum_{x \in \mathcal{I}_i} \log \bar{z}^i_{x, m_i+1} - \sum_{y \in \mathcal{J}_i} \log \bar{z}^i_{n_i+1, y},
\end{equation}
The final loss is computed by averaging the individual loss over all sampled superpoint matches: $\mathcal{L}_p = \frac{1}{N_g} \sum^{N_g}_{i=1} \mathcal{L}_{p, i}$.

%% file: figures/pipeline.tex

\begin{figure*}[t]
  \begin{overpic}[width=1.0\linewidth]{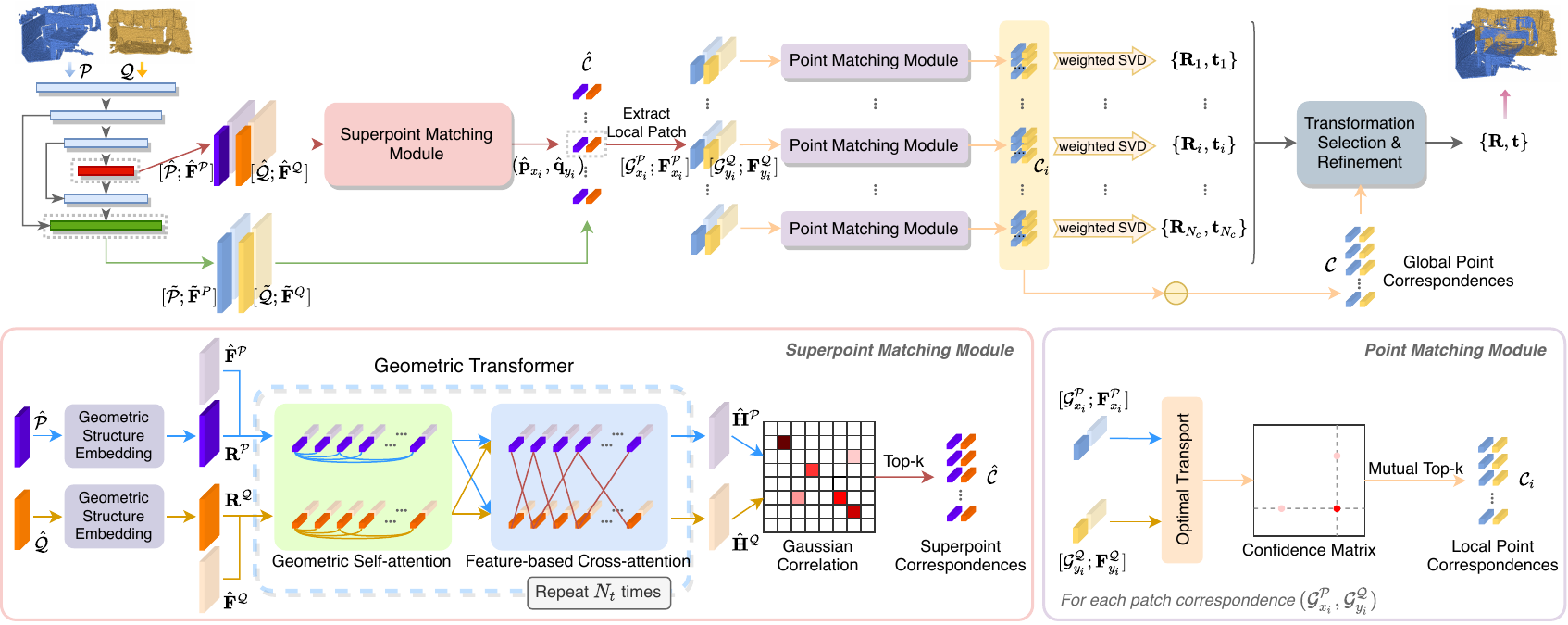}
  \put(0,40.3){\scriptsize \rule[-0.4ex]{0.2ex}{1.0em}\rule[0.6ex]{2.2ex}{0.1em} 1. Feature Extraction \rule[0.6ex]{2.2ex}{0.1em}\rule[-0.4ex]{0.2ex}{1.0em}}
  \put(15.55,40.3){\scriptsize \rule[-0.4ex]{0.2ex}{1.0em}\rule[0.6ex]{6ex}{0.1em} 2. Superpoint Matching \rule[0.6ex]{6ex}{0.1em}\rule[-0.4ex]{0.2ex}{1.0em}}
  \put(37.4,40.3){\scriptsize \rule[-0.4ex]{0.2ex}{1.0em}\rule[0.6ex]{13.2ex}{0.1em} 3. Point Matching \rule[0.6ex]{13.2ex}{0.1em}\rule[-0.4ex]{0.2ex}{1.0em}}
  \put(65.2,40.3){\scriptsize \rule[-0.4ex]{0.2ex}{1.0em}\rule[0.6ex]{12.7ex}{0.1em} 4. Local-to-Global Registration \rule[0.6ex]{12.7ex}{0.1em}\rule[-0.4ex]{0.2ex}{1.0em}}
  \end{overpic}
  \vspace{-20pt}
  \caption{The backbone downsamples the input point clouds and learns features in multiple resolution levels. The Superpoint Matching Module extracts high-quality superpoint correspondences between $\hat{\mathcal{P}}$ and $\hat{\mathcal{Q}}$ using the Geometric Transformer which iteratively encodes intra-point-cloud geometric structures and inter-point-cloud geometric consistency. The superpoint correspondences are then propagated to dense points $\tilde{\mathcal{P}}$ and $\tilde{\mathcal{Q}}$ by the Point Matching Module. Finally, the transformation is computed with a local-to-global registration method.}
  \label{fig:overview}
  \vspace{-10pt}
\end{figure*}

%% file: figures/geotr.tex

\begin{figure}[t]
  \centering
  \begin{overpic}[width=1.0\linewidth]{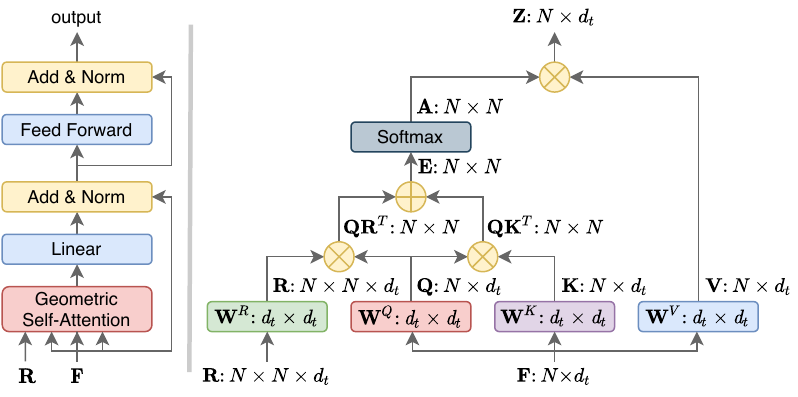}
  \end{overpic}
  \vspace{-20pt}
  \caption{Left: The structure of geometric self-attention module. Right: The computation graph of geometric self-attention.}
  \label{fig:geotr}
  \vspace{-10pt}
\end{figure}

%% file: figures/rge.tex

\begin{figure}[t]
  \begin{overpic}[width=1.0\linewidth]{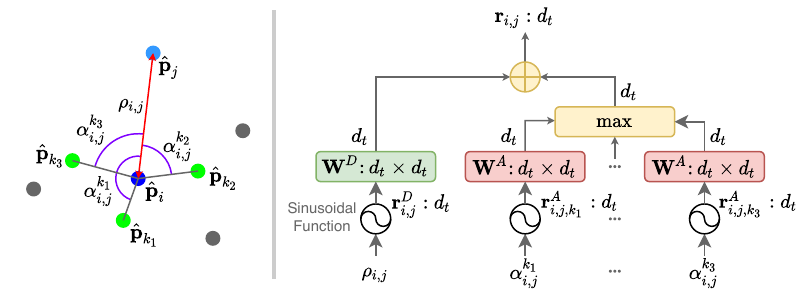}
  \end{overpic}
  \vspace{-20pt}
  \caption{An illustration of the distance-and-angle-based geometric structure encoding and its computation.}
  \label{fig:rge}
  \vspace{-10pt}
\end{figure}

%% file: results.tex

\section{Experiments}
\label{sec:experiments}

We evaluate GeoTransformer on indoor 3DMatch~\cite{zeng20173dmatch} and 3DLoMatch~\cite{huang2021predator} benchmarks (\cref{sec:exp-indoor}) and outdoor KITTI odometry~\cite{geiger2012we} benchmark (\cref{sec:exp-outdoor}).
We introduce the implementation details in Appx.~\textcolor{red}{A.3}.




\subsection{Indoor Benchmarks: 3DMatch \& 3DLoMatch}
\label{sec:exp-indoor}

\ptitle{Dataset.}
3DMatch~\cite{zeng20173dmatch} contains $62$ scenes among which $46$ are used for training, $8$ for validation and $8$ for testing.
We use the training data preprocessed by~\cite{huang2021predator} and evaluate on both 3DMatch and 3DLoMatch~\cite{huang2021predator} protocols.
The point cloud pairs in 3DMatch have $>30\%$ overlap, while those in 3DLoMatch have low overlap of $10\%$ $\sim$ $30\%$.

\ptitle{Metrics.}
Following~\cite{bai2020d3feat,huang2021predator}, we evaluate the performance with three metrics:
(1) \emph{Inlier Ratio} (IR), the fraction of putative correspondences whose residuals are below a certain threshold (\ie, $0.1\text{m}$) under the ground-truth transformation,
(2) \emph{Feature Matching Recall} (FMR), the fraction of point cloud pairs whose inlier ratio is above a certain threshold (\ie, $5\%$), and
(3) \emph{Registration Recall} (RR), the fraction of point cloud pairs whose transformation error is smaller than a certain threshold (\ie, $\mathrm{RMSE} < 0.2\text{m}$).

\begin{table}[!t]
\setlength{\tabcolsep}{1.8pt}
\scriptsize
\centering
\begin{tabular}{l|ccccc|ccccc}
\toprule
 & \multicolumn{5}{c|}{3DMatch} & \multicolumn{5}{c}{3DLoMatch} \\
\# Samples & 5000 & 2500 & 1000 & 500 & 250 & 5000 & 2500 & 1000 & 500 & 250 \\
\midrule
\multicolumn{11}{c}{\emph{Feature Matching Recall} (\%) $\uparrow$} \\
\midrule
PerfectMatch~\cite{gojcic2019perfect} & 95.0 & 94.3 & 92.9 & 90.1 & 82.9 & 63.6 & 61.7 & 53.6 & 45.2 & 34.2 \\
FCGF~\cite{choy2019fully} & 97.4 & 97.3 & 97.0 & 96.7 & 96.6 & 76.6 & 75.4 & 74.2 & 71.7 & 67.3 \\
D3Feat~\cite{bai2020d3feat} & 95.6 & 95.4 & 94.5 & 94.1 & 93.1 & 67.3 & 66.7 & 67.0 & 66.7 & 66.5 \\
SpinNet~\cite{ao2021spinnet} & 97.6 & 97.2 & 96.8 & 95.5 & 94.3 & 75.3 & 74.9 & 72.5 & 70.0 & 63.6 \\
Predator~\cite{huang2021predator} & 96.6 & 96.6 & 96.5 & 96.3 & 96.5 & 78.6 & 77.4 & 76.3 & 75.7 & 75.3 \\
YOHO~\cite{wang2021you} & \textbf{98.2} & 97.6 & 97.5 & 97.7 & 96.0 & 79.4 & 78.1 & 76.3 & 73.8 & 69.1 \\
CoFiNet~\cite{yu2021cofinet} & \underline{98.1} & \textbf{98.3} & \textbf{98.1} & \textbf{98.2} & \textbf{98.3} & \underline{83.1} & \underline{83.5} & \underline{83.3} & \underline{83.1} & \underline{82.6} \\
GeoTransformer (\emph{ours}) & 97.9 & \underline{97.9} & \underline{97.9} & \underline{97.9} & \underline{97.6} & \textbf{88.3} & \textbf{88.6} & \textbf{88.8} & \textbf{88.6} & \textbf{88.3} \\ 
\midrule
\multicolumn{11}{c}{\emph{Inlier Ratio} (\%) $\uparrow$} \\
\midrule
PerfectMatch~\cite{gojcic2019perfect} & 36.0 & 32.5 & 26.4 & 21.5 & 16.4 & 11.4 & 10.1 & 8.0 & 6.4 & 4.8 \\
FCGF~\cite{choy2019fully} & 56.8 & 54.1 & 48.7 & 42.5 & 34.1 & 21.4 & 20.0 & 17.2 & 14.8 & 11.6 \\
D3Feat~\cite{bai2020d3feat} & 39.0 & 38.8 & 40.4 & 41.5 & 41.8 & 13.2 & 13.1 & 14.0 & 14.6 & 15.0 \\
SpinNet~\cite{ao2021spinnet} & 47.5 & 44.7 & 39.4 & 33.9 & 27.6 & 20.5 & 19.0 & 16.3 & 13.8 & 11.1 \\
Predator~\cite{huang2021predator} & 58.0 & 58.4 & \underline{57.1} & \underline{54.1} & 49.3 & \underline{26.7} & \underline{28.1} & \underline{28.3} & \underline{27.5} & 25.8 \\
YOHO~\cite{wang2021you} & \underline{64.4} & \underline{60.7} & 55.7 & 46.4 & 41.2 & 25.9 & 23.3 & 22.6 & 18.2 & 15.0 \\
CoFiNet~\cite{yu2021cofinet} & 49.8 & 51.2 & 51.9 & 52.2 & \underline{52.2} & 24.4 & 25.9 & 26.7 & 26.8 & \underline{26.9} \\
GeoTransformer (\emph{ours}) & \textbf{71.9} & \textbf{75.2} & \textbf{76.0} & \textbf{82.2} & \textbf{85.1} & \textbf{43.5} & \textbf{45.3} & \textbf{46.2} & \textbf{52.9} & \textbf{57.7} \\
\midrule
\multicolumn{11}{c}{\emph{Registration Recall} (\%) $\uparrow$} \\
\midrule
PerfectMatch~\cite{gojcic2019perfect} & 78.4 & 76.2 & 71.4 & 67.6 & 50.8 & 33.0 & 29.0 & 23.3 & 17.0 & 11.0 \\
FCGF~\cite{choy2019fully} & 85.1 & 84.7 & 83.3 & 81.6 & 71.4 & 40.1 & 41.7 & 38.2 & 35.4 & 26.8  \\
D3Feat~\cite{bai2020d3feat} & 81.6 & 84.5 & 83.4 & 82.4 & 77.9 & 37.2 & 42.7 & 46.9 & 43.8 & 39.1 \\
SpinNet~\cite{ao2021spinnet} & 88.6 & 86.6 & 85.5 & 83.5 & 70.2 & 59.8 & 54.9 & 48.3 & 39.8 & 26.8 \\
Predator~\cite{huang2021predator} & 89.0 & 89.9 & \underline{90.6} & 88.5 & 86.6 & 59.8 & 61.2 & 62.4 & 60.8 & 58.1 \\
YOHO~\cite{wang2021you} & \underline{90.8} & \underline{90.3} & 89.1 & \underline{88.6} & 84.5 & 65.2 & 65.5 & 63.2 & 56.5 & 48.0 \\
CoFiNet~\cite{yu2021cofinet} & 89.3 & 88.9 & 88.4 & 87.4 & \underline{87.0} & \underline{67.5} & \underline{66.2} & \underline{64.2} & \underline{63.1} & \underline{61.0} \\
GeoTransformer (\emph{ours}) & \textbf{92.0} & \textbf{91.8} & \textbf{91.8} & \textbf{91.4} & \textbf{91.2} & \textbf{75.0} & \textbf{74.8} & \textbf{74.2} & \textbf{74.1} & \textbf{73.5} \\
\bottomrule
\end{tabular}
\vspace{-5pt}
\caption{
Evaluation results on 3DMatch and 3DLoMatch.
The comparison with deep robust estimators is present in Appx.~\textcolor{red}{D.2}.
}
\label{table:results-3dmatch}
\vspace{-10pt}
\end{table}

\ptitle{Correspondence results.}
We first compare the correspondence results of our method with the recent state of the arts: PerfectMatch~\cite{gojcic2019perfect}, FCGF~\cite{choy2019fully}, D3Feat~\cite{bai2020d3feat}, SpinNet~\cite{ao2021spinnet}, Predator~\cite{huang2021predator}, YOHO~\cite{wang2021you} and CoFiNet~\cite{yu2021cofinet} in \cref{table:results-3dmatch}(top and middle).
Following~\cite{bai2020d3feat,huang2021predator}, we report the results with different numbers of correspondences.
The details of the correspondence sampling schemes are given in Appx.~\textcolor{red}{A.3}.
For \emph{Feature Matching Recall}, our method achieves improvements of at least $5$ percentage points (pp) on 3DLoMatch, demonstrating its effectiveness in low-overlap cases.
For \emph{Inlier Ratio}, the improvements are even more prominent.
It surpasses the baselines consistently by $7{\sim}33$ pp on 3DMatch and $17{\sim}31$ pp on 3DLoMatch.
The gain is larger with less correspondences.
It implies that our method extracts more reliable correspondences.

\ptitle{Registration results.}
\label{sec:exp-registration}
To evaluate the registration performance, we first compare the \emph{Registration Recall} obtained by RANSAC in \cref{table:results-3dmatch}(bottom).
Following~\cite{bai2020d3feat,huang2021predator}, we run $50$K RANSAC iterations to estimate the transformation.
GeoTransformer attains new state-of-the-art results on both 3DMatch and 3DLoMatch.
It outperforms the previous best by $1.2$ pp
on 3DMatch and $7.5$ pp
on 3DLoMatch, showing its efficacy in both high- and low-overlap scenarios.
More importantly, our method is quite stable under different numbers of samples, so it does not require sampling a large number of correspondences to boost the performance as previous methods \cite{choy2019fully,ao2021spinnet,wang2021you,yu2021cofinet}.

\begin{table}[!t]
\scriptsize
\setlength{\tabcolsep}{1pt}
\centering
\begin{tabular}{l|c|c|cc|ccc}
\toprule
\multirow{2}{*}{Model} & \multirow{2}{*}{Estimator} & \multirow{2}{*}{\#Samples} & \multicolumn{2}{c|}{RR(\%)} & \multicolumn{3}{c}{Time(s)} \\
 & & & 3DM & 3DLM & Model & Pose & Total\\
\midrule
FCGF~\cite{choy2019fully} & RANSAC-\emph{50k} & 5000 & 85.1 & 40.1 & 0.052 & 3.326 & 3.378 \\
D3Feat~\cite{bai2020d3feat} & RANSAC-\emph{50k} & 5000 & 81.6 & 37.2 & 0.024 & 3.088 & 3.112 \\
SpinNet~\cite{ao2021spinnet} & RANSAC-\emph{50k} & 5000 & 88.6 & 59.8 & 60.248 & 0.388 & 60.636 \\
Predator~\cite{huang2021predator} & RANSAC-\emph{50k} & 5000 & 89.0 & 59.8 & 0.032 & 5.120 & 5.152 \\
CoFiNet~\cite{yu2021cofinet} & RANSAC-\emph{50k} & 5000 & \underline{89.3} & \underline{67.5} & 0.115 & 1.807 & 1.922 \\
GeoTransformer (\emph{ours}) & RANSAC-\emph{50k} & 5000 & \textbf{92.0} & \textbf{75.0} & 0.075 & 1.558 & 1.633 \\
\midrule
FCGF~\cite{choy2019fully} & weighted SVD & 250 & 42.1 & 3.9 & 0.052 & 0.008 & 0.056 \\  
D3Feat~\cite{bai2020d3feat} & weighted SVD & 250 & 37.4 & 2.8 & 0.024 & 0.008 & 0.032 \\  
SpinNet~\cite{ao2021spinnet} & weighted SVD & 250 & 34.0 & 2.5 & 60.248 & 0.006 & 60.254 \\  
Predator~\cite{huang2021predator} & weighted SVD & 250 & 50.0 & 6.4 & 0.032 & 0.009 & 0.041 \\  
CoFiNet~\cite{yu2021cofinet} & weighted SVD & 250 & \underline{64.6} & \underline{21.6} & 0.115 & 0.003 & 0.118 \\
GeoTransformer (\emph{ours}) & weighted SVD & 250 & \textbf{86.5} & \textbf{59.9} & 0.075 & 0.003 & 0.078 \\
\midrule
CoFiNet~\cite{yu2021cofinet} & LGR & all & 87.6 & 64.8 & 0.115 & 0.028 & 0.143 \\
GeoTransformer (\emph{ours}) & LGR & all & \textbf{91.5} & \textbf{74.0} & 0.075 & 0.013 & 0.088 \\
\bottomrule
\end{tabular}
\vspace{-5pt}
\caption{
Registration results w/o RANSAC on 3DMatch (3DM) and 3DLoMatch (3DLM).
The \emph{model time} is the time for feature extraction, while the \emph{pose time} is for transformation estimation.
}
\label{table:direct}
\vspace{-10pt}
\end{table}

We then compare the registration results \emph{without} using RANSAC in \cref{table:direct}.
We start with weighted SVD over correspondences in solving for alignment transformation.
The baselines either fail to achieve reasonable results or suffer from severe performance degradation.
In contrast, GeoTransformer (with weighted SVD) achieves the registration recall of $86.5\%$ on 3DMatch and $59.9\%$ on 3DLoMatch, close to Predator with RANSAC.
Without outlier filtering by RANSAC, high inlier ratio is necessary for successful registration.
However, high inlier ratio does not necessarily lead to high registration recall since the correspondences could cluster together as noted in~\cite{huang2021predator}. Nevertheless, our method without RANSAC performs well by extracting reliable and well-distributed superpoint correspondences.

When using our local-to-global registration (LGR) for computing transformation, our method brings the registration recall to $91.5\%$ on 3DMatch and $74.0\%$ on 3DLoMatch, surpassing all RANSAC-based baselines by a large margin.
The results are also very close to those of ours with RANSAC, but LGR gains over $100$ times acceleration over RANSAC in the pose time.
These results demonstrate the superiority of our method in both accuracy and speed.

\input{figures/gallery}

\ptitle{Ablation studies.}
\label{sec:exp-ablation}
We conduct extensive ablation studies for a better understanding of the various modules in our method\footnote{Due to space limit, we present some ablation studies in Appx.~\textcolor{red}{D.3}.}.
To evaluate superpoint (patch) matching, we introduce another metric \emph{Patch Inlier Ratio} (PIR) which is the fraction of patch matches with actual overlap.
The FMR and IR are reported with \emph{all} dense point correspondences, with LGR being used for registration.

\begin{table}[!t]
\scriptsize
\setlength{\tabcolsep}{3pt}
\centering
\begin{tabular}{l|cccc|cccc}
\toprule
 \multirow{2}{*}{Model} & \multicolumn{4}{c|}{3DMatch} & \multicolumn{4}{c}{3DLoMatch} \\
 & PIR & FMR & IR & RR & PIR & FMR & IR & RR \\
\midrule
(a) graph neural network & 73.3 & 97.9 & 56.5 & 89.5 & 39.4 & 84.9 & 29.2 & 69.8 \\
(b) vanilla self-attention & 79.6 & 97.9 & 60.1 & 89.0 & 45.2 & 85.6 & 32.6 & 68.4 \\
(c) self-attention w/ ACE & 83.2 & \textbf{98.1} & 68.5 & 89.3 & 48.2 & 84.3 & 38.9 & 69.3 \\
(d) self-attention w/ RCE & 80.0 & 97.9 & 66.1 & 88.5 & 46.1 & 84.6 & 37.9 & 68.7 \\
(e) self-attention w/ PPF & 83.5 & 97.5 & 68.5 & 88.6 & 49.8 & 83.8 & 39.9 & 69.5 \\
(f) self-attention w/ RDE & \underline{84.9} & \underline{98.0} & \underline{69.1} & \underline{90.7} & \underline{50.6} & \underline{85.8} & \underline{40.3} & \underline{72.1} \\
(g) geometric self-attention & \textbf{86.1} & 97.7 & \textbf{70.3} & \textbf{91.5} & \textbf{54.9} & \textbf{88.1} & \textbf{43.3} & \textbf{74.0} \\
\bottomrule
\end{tabular}
\vspace{-5pt}
\caption{
Ablation experiments of the geometric self-attention.
}
\label{table:ablation-study-rge}
\vspace{-10pt}
\end{table}

\begin{table}[!t]
\scriptsize
\setlength{\tabcolsep}{2.5pt}
\centering
\begin{tabular}{l|cccc|cccc}
\toprule
\multirow{2}{*}{Model} & \multicolumn{4}{c|}{3DMatch} & \multicolumn{4}{c}{3DLoMatch} \\
 & PIR & FMR & IR & RR & PIR & FMR & IR & RR \\
\midrule
(a) cross-entropy loss & 80.0 & 97.7 & 65.7 & 90.0 & 45.9 & 85.1 & 37.4 & 68.4 \\
(b) weighted cross-entropy loss & 83.2 & \textbf{98.0} & 67.4 & 90.0 & 49.0 & \underline{86.2} & 38.6 & 70.7 \\
(c) circle loss & \underline{85.1} & \underline{97.8} & \underline{69.5} & \underline{90.4} & \underline{51.5} & 86.1 & \underline{41.3} & \underline{71.5} \\
(d) overlap-aware circle loss & \textbf{86.1} & 97.7 & \textbf{70.3} & \textbf{91.5} & \textbf{54.9} & \textbf{88.1} & \textbf{43.3} & \textbf{74.0} \\
\bottomrule
\end{tabular}
\vspace{-5pt}
\caption{
Ablation experiments of the overlap-aware circle loss.
}
\label{table:ablation-study-ocl}
\vspace{-10pt}
\end{table}

To study the effectiveness of the \emph{geometric self-attention}, we compare seven methods for intra-point-cloud feature learning in \cref{table:ablation-study-rge}:
(a) graph neural network~\cite{huang2021predator}, (b) self-attention with no positional embedding~\cite{yu2021cofinet}, (c) absolute coordinate embedding~\cite{sarlin2020superglue}, (d) relative coordinate embedding~\cite{zhao2021point}, (e) point pair features~\cite{drost2010model,raposo2017using} embedding, (f) pair-wise distance embedding, (g) geometric structure embedding.
Generally, injecting geometric information boosts the performance.
But the gains of coordinate-based embeddings are limited due to their transformation variance.
Surprisingly, GNN performs well on RR thanks to the transformation invariance of $k$NN graphs.
However, it suffers from limited receptive fields which harms the IR performance.
Although PPF embedding is theoretically invariant to transformation, it is hard to estimate accurate normals for the superpoints in practice, which leads to inferior performance.
Our method outperforms the alternatives by a large margin on all the metrics, especially in the low-overlap scenarios, even with only the pair-wise distance embedding, demontrating the strong robustness of our method.

\input{figures/attention}

Next, we ablate the \emph{overlap-aware circle loss} in \cref{table:ablation-study-ocl}.
We compare four loss functions for supervising the superpoint matching: (a) cross-entropy loss~\cite{sarlin2020superglue}, (b) weighted cross-entropy loss~\cite{yu2021cofinet}, (c) circle loss~\cite{sun2020circle}, and (d) overlap-aware circle loss.
For the first two models, an optimal transport layer is used to compute the matching matrix as in~\cite{yu2021cofinet}.
Circle loss works much better than the two variants of cross-entropy loss, verifying the effectiveness of supervising superpoint matching in a metric learning fashion.
Our overlap-aware circle loss beats the vanilla circle loss by a large margin on all the metrics.

\ptitle{Qualitative results.}
\cref{fig:gallery} provides a gallery of the registration results of the models with vanilla self-attention and our geometric self-attention.
Geometric self-attention helps infer patch matches in structure-less regions from their geometric relationships to more salient regions ($1^{\text{st}}$ row) and reject outlier matches which are similar in the feature space but different in positions ($2^{\text{nd}}$ and $3^{\text{rd}}$ rows).

\cref{fig:attention} visualizes the attention scores learned by our geometric self-attention, which exhibits significant consistency between the anchor patch matches.
It shows that our method is able to learn inter-point-cloud geometric consistency which is important to accurate correspondences.

\subsection{Outdoor Benchmark: KITTI odometry}
\label{sec:exp-outdoor}

\ptitle{Dataset.}
KITTI odometry~\cite{geiger2012we} consists of 11 sequences of outdoor driving scenarios scanned by LiDAR.
We follow~\cite{choy2019fully,bai2020d3feat,huang2021predator} and use sequences 0-5 for training, 6-7 for validation and 8-10 for testing.
As in~\cite{choy2019fully,bai2020d3feat,huang2021predator}, the ground-truth poses are refined with ICP and we only use point cloud pairs that are at least $10\text{m}$ away for evaluation.

\ptitle{Metrics.}
We follow~\cite{huang2021predator} to evaluate our GeoTransformer with three metrics:
(1) \emph{Relative Rotation Error} (RRE), the geodesic distance between estimated and ground-truth rotation matrices,
(2) \emph{Relative Translation Error} (RTE), the Euclidean distance between estimated and ground-truth translation vectors, and
(3) \emph{Registration Recall} (RR), the fraction of point cloud pairs whose RRE and RTE are both below certain thresholds (\ie, RRE$<$5$^\circ$ and RTE$<$2m).

\begin{table}[!t]
\scriptsize
\centering
\begin{tabular}{l|ccc}
\toprule
Model & RTE(cm) & RRE($^{\circ}$) & RR(\%) \\
\midrule
3DFeat-Net~\cite{yew20183dfeat} & 25.9 & \textbf{0.25} & 96.0 \\
FCGF~\cite{choy2019fully} & 9.5 & 0.30 & \underline{96.6} \\
D3Feat~\cite{bai2020d3feat} & \underline{7.2} & 0.30 & \textbf{99.8} \\
SpinNet~\cite{ao2021spinnet} & 9.9 & 0.47 & 99.1 \\
Predator~\cite{huang2021predator} & \textbf{6.8} & \underline{0.27} & \textbf{99.8} \\
CoFiNet~\cite{yu2021cofinet} & 8.2 & 0.41 & \textbf{99.8} \\
GeoTransformer (\emph{ours}, RANSAC-\emph{50k}) & 7.4 & \underline{0.27} & \textbf{99.8} \\
\midrule
FMR~\cite{huang2020feature} & $\sim$66 & 1.49 & 90.6 \\
DGR~\cite{choy2020deep} & $\sim$32 & 0.37 & 98.7 \\
HRegNet~\cite{lu2021hregnet} & $\sim$\underline{12} & \underline{0.29} & \underline{99.7} \\
GeoTransformer (\emph{ours}, LGR) & \textbf{6.8} & \textbf{0.24} & \textbf{99.8} \\
\bottomrule
\end{tabular}
\vspace{-5pt}
\caption{
Registration results on KITTI odometry.
The comparison with deep robust estimators is present in Appx.~\textcolor{red}{D.2}.
}
\vspace{-10pt}
\label{table:kitti}
\end{table}

\ptitle{Registration results.}
In \cref{table:kitti}(top), we compare to the state-of-the-art \emph{RANSAC-based} methods: 3DFeat-Net~\cite{yew20183dfeat}, FCGF~\cite{choy2019fully}, D3Feat~\cite{bai2020d3feat}, SpinNet~\cite{ao2021spinnet}, Predator~\cite{huang2021predator} and CoFiNet~\cite{yu2021cofinet}.
Our method performs on par with these methods, showing good generality on outdoor scenes.
We further compare to three \emph{RANSAC-free} methods in \cref{table:kitti}(bottom): FMR~\cite{huang2020feature}, DGR~\cite{choy2020deep} and HRegNet~\cite{lu2021hregnet}.
Our method outperforms all the baselines by large margin.
In addition, our method with LGR beats all the RANSAC-based methods.

%% file: figures/gallery.tex

\begin{figure*}[t]
  \begin{overpic}[width=1.0\linewidth]{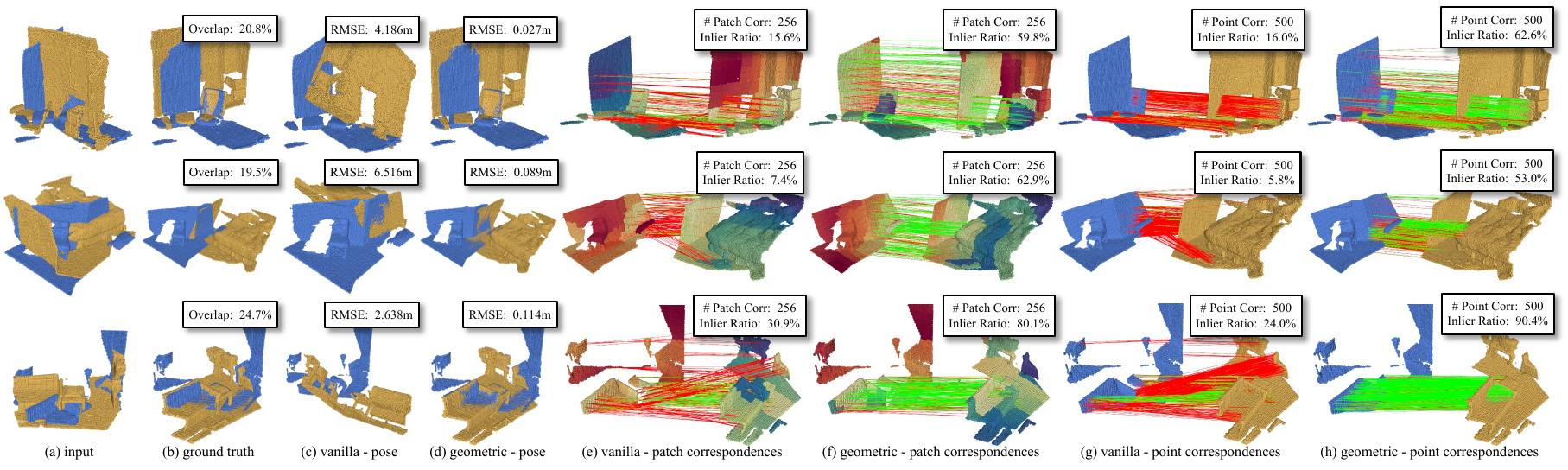}
  \end{overpic}
  \vspace{-20pt}
  \caption{Registration results of the models with vanilla self-attention and geometric self-attention. In the columns (e) and (f), we visualize the features of the patches with t-SNE. In the first row, the geometric self-attention helps find the inlier matches on the structure-less wall based on their geometric relationships to the more salient regions (\eg, the chairs). In the following rows, the geometric self-attention helps reject the outlier matches between the similar flat or corner patches based on their geometric relationships to the bed or the sofa.
  }
  \label{fig:gallery}
  \vspace{-10pt}
\end{figure*}


%% file: figures/attention.tex

\begin{figure}[t]
  \begin{overpic}[width=1.0\linewidth]{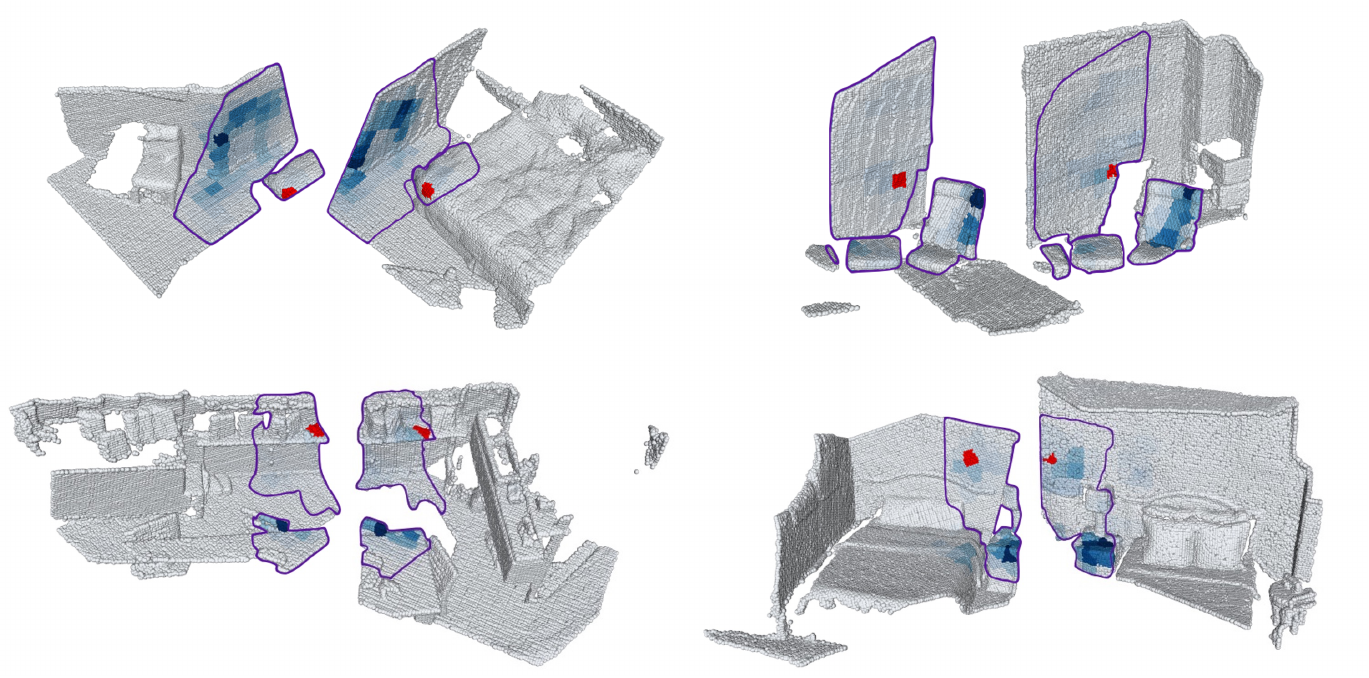}
  \end{overpic}
  \vspace{-15pt}
  \caption{Visualizing geometric self-attention scores on four pairs of point clouds. The overlap areas are delineated with {\color[rgb]{0.35,0.09,0.60}purple} lines. The anchor patches (in correspondence) are highlighted in {\color{red}red} and the attention scores to other patches are color-coded (\emph{deeper is larger}). Note how the attention patterns of the two matching anchors are consistent even across disjoint overlap areas.}
  \label{fig:attention}
  \vspace{-10pt}
\end{figure}

%% file: conclusion.tex

\section{Conclusion}

We have presented Geometric Transformer to learn robust coarse-to-fine correspondence for point cloud registration. Through encoding pair-wise distances and triplet-wise angles among superpoints, our method captures the geometric consistency across point clouds with transformation invariance.
Thanks to the reliable correspondences, it attains fast and accurate registration in a RANSAC-free manner.
We discuss the limitations of our method in Appx.~\textcolor{red}{E}.
In the future, we would like to extend our method to cross-modality (\eg, 2D-3D) registration with richer applications.




\ptitle{Acknowledgement.} This work is supported in part by the NSFC (62132021, 62102435) and the National Key R\&D Program of China (2018AAA0102200).


%% file: appendix.tex

\begin{appendix}


\section{Network Architecture Details}

\subsection{Geometric Structure Embedding}
\label{supp:rge}

First, we provide the detailed computation for our geometric structure embedding.
The geometric structure embedding encodes distances in superpoint pairs and angles in superpoint triplets.
Due to the continuity of the sinusoidal embedding function~\cite{vaswani2017attention}, we use it instead of learned embedding vectors to compute the pair-wise distance embedding and the triplet-wise angular embedding.

Given the distance $\rho_{i, j} \tight{=} \lVert \hat{\textbf{p}}_i - \hat{\textbf{p}}_j \rVert_2$ between $\hat{\textbf{p}}_i$ and $\hat{\textbf{p}}_j$, the pair-wise distance embedding $\textbf{r}^D_{i, j}$ is computed as:
\begin{equation}
\left\{
\begin{aligned}
r^D_{i, j, 2k} & = \sin(\frac{d_{i, j} / \sigma_d}{10000^{2k / d_t}}) \\
r^D_{i, j, 2k+1} & = \cos(\frac{d_{i, j} / \sigma_d}{10000^{2k / d_t}})
\end{aligned},
\right.
\end{equation}
where $d_t$ is the feature dimension, and $\sigma_d$ is a temperature which controls the sensitivity to distance variations.

The triplet-wise angular embedding can be computed in the same way.
Given the angle $\alpha^k_{i, j}$, the triplet-wise angular embedding $\textbf{r}^A_{i, j, k}$ is computed as:
\begin{equation}
\left\{
\begin{aligned}
r^A_{i, j, k, 2x} & = \sin(\frac{\alpha^k_{i, j} / \sigma_a}{10000^{2x / d_t}}) \\
r^A_{i, j, k, 2x+1} & = \cos(\frac{\alpha^k_{i, j} / \sigma_a}{10000^{2x / d_t}})
\end{aligned},
\right.
\end{equation}
where $\sigma_a$ is another temperature to control the sensitivity to angular variantions.

\subsection{Point Matching Module}

For completeness, we then provide the details of the optimal transport layer~\cite{sarlin2020superglue} in the point matching module.
For each superpoint correspondence $(\hat{\textbf{p}}_{x_i}, \hat{\textbf{q}}_{y_i})$, its local point correspondences are extracted from their local patches $\mathcal{G}^{\mathcal{P}}_{x_i}$ and $\mathcal{G}^{\mathcal{Q}}_{y_i}$.
We first compute a cost matrix $\textbf{C}_i \tight{\in} \mathbb{R}^{n_i \times m_i}$ using the feature matrices of the two patches:
\begin{equation}
\textbf{C}_i = \textbf{F}^{\mathcal{P}}_{x_i} (\textbf{F}^{\mathcal{Q}}_{y_i})^T / \sqrt{\tilde{d}},
\end{equation}
where $n_i = \lvert \mathcal{G}^{\mathcal{P}}_{x_i} \rvert$, $m_i = \lvert \mathcal{G}^{\mathcal{Q}}_{y_i} \rvert$.
The cost matrix $\textbf{C}_i$ is then augmented to $\bar{\textbf{C}}_i \medium{\in} \mathcal{R}^{(n_i + 1) \times (m_i + 1)}$ by appending a new row and a new column filled with a learnable dustbin parameter $\alpha$ as in~\cite{sarlin2020superglue}.
The point matching problem can then be formulated as an optimal transport problem which maximizes $\sum_{j, k} \bar{\textbf{C}}_i \cdot \bar{\textbf{Z}}_i$, where $\bar{\textbf{Z}}_i \medium{\in} \mathcal{R}^{(n_i + 1) \times (m_i + 1)}$ is the soft assignment matrix satisfying:
\begin{align}
\sum_{k=1}^{m_i + 1} \bar{z}^i_{j, k} = \begin{cases}
1, & 1 \leq j \leq n_i \\
m_i, & j = n_i + 1
\end{cases}, \\
\sum_{j=1}^{n_i + 1} \bar{z}^i_{j, k} = \begin{cases}
1, & 1 \leq k \leq m_i \\
n_i, & k = m_i + 1
\end{cases}.
\end{align}
Here $\bar{\textbf{Z}}_i$ can be solved by the differentiable Sinkhorn algorithm~\cite{sinkhorn1967concerning} with doubly-normalization iterations:
\begin{align}
{}^{(t)}u^i_j & = \log \alpha^i_j - \log \sum_{k=1}^{m_i+1} \exp(\bar{c}^{\hspace{1pt}i}_{j, k} + {}^{(t-1)}v^i_k), \\
{}^{(t)}v^i_k & = \log \beta^i_k - \log \sum_{j=1}^{n_i+1} \exp(\bar{c}^{\hspace{1pt}i}_{j, k} + {}^{(t)}u^i_j),
\end{align}
where
\begin{align}
\alpha^i_j & = \begin{cases}
\frac{1}{n_i + m_i}, & 1 \leq j \leq n_i \\
\frac{m_i}{n_i + m_i}, & j = n_i + 1
\end{cases}, \\
\beta^i_k & = \begin{cases}
\frac{1}{n_i + m_i}, & 1 \leq k \leq m_i \\
\frac{n_i}{n_i + m_i}, & k = m_i + 1
\end{cases}.
\end{align}
The algorithm starts with ${}^{(0)}\textbf{u} \medium{=} \textbf{0}^{n_i + 1}$ and ${}^{(0)}\textbf{v} \medium{=} \textbf{0}^{m_i + 1}$. The assignment matrix $\bar{\textbf{Z}}_i$ is then computed as:
\begin{equation}
\bar{z}^{\hspace{1pt}i}_{j, k} \medium{=} \exp(\bar{c}^{\hspace{1pt}i}_{j, k} + u^i_j + v^i_k) \medium{\cdot} (n_i \tight{+} m_i).
\end{equation}
We run $t_0 \medium{=} 100$ Sinkhorn iterations following~\cite{sarlin2020superglue}.
$\bar{\textbf{Z}}_i$ is then recovered to $\textbf{Z}_i  \medium{\in} \mathbb{R}^{n_i \times m_i}$ by dropping the last row and the last column, which is used as the confidence matrix of the candidate matches.
The local point correspondences are extracted by mutual top-$k$ selection on $\textbf{Z}_i$.
We ignore the matches whose confidence scores are too small (\ie, $z^i_{x_j, y_j} \tight{<} 0.05$).
The hyper-parameter $k$ controls the number of point correspondences as described in~\cref{supp:details}.
At last, the final global dense point correspondences are generated by combining the local point correspondences from all superpoint correspondences together.

\subsection{Network Configurations}
\label{supp:details}

\paragraph{Backbone.}

We use a KPConv-FPN backbone for feature extraction.
The grid subsampling scheme~\cite{thomas2019kpconv} is used to downsample the point clouds.
Before being fed into the backbone, the input point clouds are first downsampled with a voxel size of $2.5\text{cm}$ on 3DMatch and $30\text{cm}$ on KITTI.
The voxel size is then doubled in each downsampling operation.
We use a 4-stage backbone for 3DMatch and a 5-stage backbone for KITTI because the point clouds in KITTI are much larger than those in 3DMatch.
The configurations of KPConv are the same as in~\cite{huang2021predator}.
And we use group normalization~\cite{wu2018group} with $8$ groups after the KPConv layers.
The detailed network configurations are shown in \cref{table:architecture}.

\begin{table}[!t]
\scriptsize
\setlength{\tabcolsep}{5pt}
\centering
\begin{tabular}{c|c|c}
\toprule
Stage & 3DMatch & KITTI \\
\midrule
\multicolumn{3}{c}{\emph{Backbone}} \\
\midrule
\multirow{2}{*}{1} & KPConv($1 \tight{\rightarrow} 64$) & KPConv($1 \tight{\rightarrow} 64$) \\
 & ResBlock($64 \rightarrow 128$) & ResBlock($64 \rightarrow 128$) \\
\midrule
\multirow{3}{*}{2} & ResBlock($64 \rightarrow 128$, strided) & ResBlock($64 \rightarrow 128$, strided) \\
 & ResBlock($128 \rightarrow 256$) & ResBlock($128 \rightarrow 256$) \\
 & ResBlock($256 \rightarrow 256$) & ResBlock($256 \rightarrow 256$) \\
\midrule
\multirow{3}{*}{3} & ResBlock($256 \rightarrow 256$, strided) & ResBlock($256 \rightarrow 256$, strided) \\
 & ResBlock($256 \rightarrow 512$) & ResBlock($256 \rightarrow 512$) \\
 & ResBlock($512 \rightarrow 512$) & ResBlock($512 \rightarrow 512$) \\
\midrule
\multirow{3}{*}{4} & ResBlock($512 \rightarrow 512$, strided) & ResBlock($512 \rightarrow 512$, strided) \\
 & ResBlock($512 \rightarrow 1024$) & ResBlock($512 \rightarrow 1024$) \\
 & ResBlock($1024 \rightarrow 1024$) & ResBlock($1024 \rightarrow 1024$) \\
\midrule
\multirow{3}{*}{5} & \multirow{3}{*}{-} & ResBlock($1024 \rightarrow 1024$, strided) \\
 & & ResBlock($1024 \rightarrow 2048$) \\
 & & ResBlock($2048 \rightarrow 2048$) \\
\midrule
\multirow{2}{*}{6} & \multirow{2}{*}{-} & NearestUpsampling \\
 & & UnaryConv($3072 \rightarrow 1024$) \\
\midrule
\multirow{2}{*}{7} & NearestUpsampling & NearestUpsampling \\
 & UnaryConv($1536 \rightarrow 512$) & UnaryConv($1536 \rightarrow 512$) \\
\midrule
\multirow{2}{*}{8} & NearestUpsampling & NearestUpsampling \\
 & UnaryConv($768 \rightarrow 256$) & UnaryConv($768 \rightarrow 256$) \\
\midrule
\multicolumn{3}{c}{\emph{Superpoint Matching Module}} \\
\midrule
1 & Linear($1024 \rightarrow 256$) & Linear($2048 \rightarrow 128$) \\
\midrule
\multirow{2}{*}{2} & GeometricSelfAttention(256, 4) & GeometricSelfAttention(128, 4) \\
 & FeatureCrossAttention(256, 4) & FeatureCrossAttention(128, 4) \\
\midrule
\multirow{2}{*}{3} & GeometricSelfAttention(256, 4) & GeometricSelfAttention(128, 4) \\
 & FeatureCrossAttention(256, 4) & FeatureCrossAttention(128, 4) \\
\midrule
\multirow{2}{*}{4} & GeometricSelfAttention(256, 4) & GeometricSelfAttention(128, 4) \\
 & FeatureCrossAttention(256, 4) & FeatureCrossAttention(128, 4) \\
\midrule
5 & Linear($256 \rightarrow 256$) & Linear($128 \rightarrow 256$) \\
\bottomrule
\end{tabular}
\caption{
Network architecture for 3DMatch and KITTI.
}
\label{table:architecture}
\end{table}

\vspace{-10pt}
\paragraph{Superpoint Matching Module.}

At the beginning of the superpoint matching module, a linear projection is used to compress the feature dimension.
For 3DMatch, the feature dimension is $256$.
For KITTI, we halve the feature dimension to $128$ to reduce memory footprint.
We then interleave the geometric self-attention module and the feature-based cross-attention module for $N_t = 3$ times:
\begin{align}
{}^{(1)}\hat{\textbf{F}}{}^{\mathcal{P}}_{\text{self}} & = \mathrm{GeometricSelfAtt}(\hat{\mathcal{P}}, \hat{\textbf{F}}{}^{\mathcal{P}} \textbf{W}_{\text{in}}), \\
{}^{(1)}\hat{\textbf{F}}{}^{\mathcal{Q}}_{\text{self}} & = \mathrm{GeometricSelfAtt}(\hat{\mathcal{Q}}, \hat{\textbf{F}}{}^{\mathcal{Q}} \textbf{W}_{\text{in}}), \\
{}^{(t)}\hat{\textbf{F}}{}^{\mathcal{P}}_{\text{self}} & = \mathrm{GeometricSelfAtt}(\hat{\mathcal{P}}, {}^{(t-1)}\hat{\textbf{F}}{}^{\mathcal{P}}_{\text{cross}}), \\
{}^{(t)}\hat{\textbf{F}}{}^{\mathcal{Q}}_{\text{self}} & = \mathrm{GeometricSelfAtt}(\hat{\mathcal{Q}}, {}^{(t-1)}\hat{\textbf{F}}{}^{\mathcal{Q}}_{\text{cross}}), \\
{}^{(t)}\hat{\textbf{F}}{}^{\mathcal{P}}_{\text{cross}} & = \mathrm{FeatureCrossAtt}({}^{(t)}\hat{\textbf{F}}{}^{\mathcal{P}}_{\text{self}}, {}^{(t)}\hat{\textbf{F}}{}^{\mathcal{Q}}_{\text{self}}), \\
{}^{(t)}\hat{\textbf{F}}{}^{\mathcal{Q}}_{\text{cross}} & = \mathrm{FeatureCrossAtt}({}^{(t)}\hat{\textbf{F}}{}^{\mathcal{Q}}_{\text{self}}, {}^{(t)}\hat{\textbf{F}}{}^{\mathcal{P}}_{\text{cross}}).
\end{align}
All attention modules have $4$ attention heads.
In the geometric structure embedding, we use $\sigma_d \tight{=} 0.2\text{m}$ on 3DMatch and $\sigma_d \tight{=} 4.8\text{m}$ on KITTI (\ie, the voxel size in the coarsest resolution level), while $\sigma_a \tight{=} 15^{\circ}$ on both datasets.
The computation of the feature-based cross-attention for $\hat{\mathcal{P}}$ is shown in \cref{fig:supp-cross-att}.
Afterwards, we use another linear projection to project the features to $256$-d, \ie, the final $\hat{\textbf{H}}{}^{\mathcal{P}}$ and $\hat{\textbf{H}}{}^{\mathcal{Q}}$:
\begin{align}
\hat{\textbf{H}}{}^{\mathcal{P}} & = {}^{(N_t)}\hat{\textbf{F}}{}^{\mathcal{P}}_{\text{cross}} \textbf{W}_{\text{out}}, \\
\hat{\textbf{H}}{}^{\mathcal{Q}} & = {}^{(N_t)}\hat{\textbf{F}}{}^{\mathcal{Q}}_{\text{cross}} \textbf{W}_{\text{out}}.
\end{align}

\input{figures/supp-cross-att}

\vspace{-10pt}
\paragraph{Local-to-Global Registration.}

In the local-to-global registration, we only use the superpoint correspondences with at least $3$ local point correspondences to compute the transformation candidates.
To select the best transformation, the acceptance radius is $\tau_a \tight{=} 10\text{cm}$ on 3DMatch and $\tau_a \tight{=} 60\text{cm}$ on KITTI.
At last, we iteratively recompute the transformation with the surviving inlier matches for $N_r \tight{=} 5$ times, which is similar with the post-refinement process in~\cite{bai2021pointdsc}.
However, we do not change the weights of the correspondences during the refinement.
The impact of the number of iterations in the refinement is studied in \cref{supp:additional-ablation}.

\vspace{-10pt}
\paragraph{Implementation details.}

We implement and evaluate our GeoTransformer with PyTorch~\cite{paszke2019pytorch} on a Xeon Glod 5218 CPU and an NVIDIA RTX 3090 GPU.
The network is trained with Adam optimizer~\cite{kingma2014adam} for $40$ epochs on 3DMatch and $80$ epochs on KITTI.
The batch size is $1$ and the weight decay is $10^{-6}$.
The learning rate starts from $10^{-4}$ and decays exponentially by $0.05$ every epoch on 3DMatch and every $4$ epochs on KITTI.
We use the matching radius of $\tau \tight{=} 5\text{cm}$ for 3DMatch and $\tau \tight{=} 60\text{cm}$ for KITTI (\ie, the voxel size in the resolution level of $\tilde{\mathcal{P}}$ and $\tilde{\mathcal{Q}}$) to determine overlapping during the generation of both superpoint-level and point-level ground-truth matches.
The same data augmentation as in~\cite{huang2021predator} is adopted.
We randomly sample $N_g \tight{=} 128$ ground-truth superpoint matches during training, and use $N_c \tight{=} 256$ putative ones during testing.

\vspace{-10pt}
\paragraph{Correspondences sampling strategy.}

For 3DMatch, we vary the hyper-parameter $k$ in the mutual top-$k$ selection of the point matching module to control the number of the point correspondences for GeoTransformer, \ie, $k \tight{=} 1$ for $250$/$500$/$1000$ matches, $k \tight{=} 2$ for $2500$ matches, and $k \tight{=} 3$ for $5000$ matches.
And we use top-$k$ selection to sample a certain number of the correspondences instead of random sampling as in~\cite{choy2019fully,huang2021predator,yu2021cofinet}, which makes our correspondences deterministic.
For the registration with LGR (Tab.~2(bottom) of our main paper), we use $k \tight{=} 3$ to generate around $6000$ correspondences for each point cloud pair.
For the baselines, we report the results from their original papers or official models in Tab.~1 of our main paper.

For the registration with weighted SVD (Tab.~2(middle) of our main paper), the correspondences of the baselines are extracted in the following manner: we first sample $5000$ keypoints and generate the correspondences with mutual nearest neighbor selection in the feature space, and then the top $250$ correspondences with the smallest feature distances are used to compute the transformation.
The weights of the correspondences are computed as $w_i = \exp(-\lVert \textbf{f}^{\mathcal{P}}_{x_i} - \textbf{f}^{\mathcal{Q}}_{y_i} \rVert^2_2)$, where $\textbf{f}^{\mathcal{P}}_{x_i}$ and $\textbf{f}^{\mathcal{Q}}_{y_i}$ are the respective descriptors of the correspondences.
In the sampling strategies that we have tried, this scheme achieves the best registration results.

For KITTI, we use $k \tight{=} 2$ and select the top $5000$ point correspondences following~\cite{bai2020d3feat,huang2021predator}.
All other hyperparameters are the same as those in 3DMatch.

\section{Metrics}
\label{supp:metrics}

Following common practice~\cite{bai2020d3feat,huang2021predator,choy2019fully}, we use different metrics for 3DMatch and KITTI.
On 3DMatch, we report \emph{Inlier Ratio}, \emph{Feature Matching Recall} and \emph{Registration Recall}.
We also report \emph{Patch Inlier Ratio} to evaluate the superpoint (patch) correspondences.
On KITTI, we report \emph{Relative Rotation Error}, \emph{Relative Translation Error} and \emph{Registration Recall}.

\subsection{3DMatch/3DLoMatch}
\label{supp:metrics-3dmatch}

\emph{Inlier Ratio} (IR) is the fraction of inlier matches among all putative point matches.
A match is considered as an inlier if the distance between the two points is smaller than $\tau_1=10\text{cm}$ under the ground-truth transformation $\bar{\textbf{T}}_{\textbf{P} \rightarrow \textbf{Q}}$:
\begin{equation}
\mathrm{IR} = \frac{1}{\lvert \mathcal{C} \rvert} \sum_{(\textbf{p}_{x_i}, \textbf{q}_{y_i}) \in \mathcal{C}} \llbracket \lVert \bar{\textbf{T}}_{\textbf{P} \rightarrow \textbf{Q}}(\textbf{p}_{x_i}) - \textbf{q}_{y_i} \rVert_2 < \tau_1 \rrbracket,
\end{equation}
where $\llbracket \cdot \rrbracket$ is the Iversion bracket.


\emph{Feature Matching Recall} (FMR) is the fraction of point cloud pairs whose IR is above $\tau_2 = 0.05$.
FMR measures the potential success during the registration:
\begin{equation}
\mathrm{FMR} = \frac{1}{M} \sum_{i=1}^{M} \llbracket \mathrm{IR}_i > \tau_2 \rrbracket,
\end{equation}
where $M$ is the number of all point cloud pairs.


\emph{Registration Recall} (RR) is the fraction of correctly registered point cloud pairs.
Two point clouds are correctly registered if their transformation error is smaller than $0.2\text{m}$.
The transformation error is computed as the root mean square error of the ground-truth correspondences $\mathcal{C}^{*}$ after applying the estimated transformation $\textbf{T}_{\textbf{P} \rightarrow \textbf{Q}}$:
\begin{align}
\mathrm{RMSE} = & \sqrt{\frac{1}{\lvert \mathcal{C}^{*} \rvert} \sum_{(\textbf{p}^*_{x_i}, \textbf{q}^*_{y_i}) \in \mathcal{C}^{*}} \lVert \textbf{T}_{\textbf{P} \rightarrow \textbf{Q}}(\textbf{p}^*_{x_i}) - \textbf{q}^*_{y_i} \rVert_2^2}, \\
\mathrm{RR} = & \frac{1}{M} \sum_{i=1}^{M} \llbracket \mathrm{RMSE}_i < 0.2\text{m} \rrbracket.
\end{align}


\emph{Patch Inlier Ratio} (PIR) is the fraction of superpoint (patch) matches with actual overlap under the ground-truth transformation.
It reflects the quality of the putative superpoint (patch) correspondences:
\begin{equation}
\mathrm{PIR} = \frac{1}{\lvert \hat{\mathcal{C}} \rvert} \sum_{(\hat{\textbf{p}}_{x_i}, \hat{\textbf{q}}_{y_i}) \in \hat{\mathcal{C}}} \llbracket \exists \tilde{\textbf{p}} \in \mathcal{G}^{\mathcal{P}}_{x_i}, \tilde{\textbf{q}} \in \mathcal{G}^{\mathcal{Q}}_{y_i} \mathrm{\ s.t.\ } \lVert \tilde{\textbf{p}} - \tilde{\textbf{q}} \rVert_2 < \tau \rrbracket,
\end{equation}
where the matching radius is $\tau = 5\text{cm}$ as stated in~\ref{supp:details}.



\subsection{KITTI}
\label{supp:metrics-kitti}

\emph{Relative Rotation Error} (RRE) is the geodesic distance in degrees between estimated and ground-truth rotation matrices.
It measures the differences between the predicted and the ground-truth rotation matrices.
\begin{equation}
\mathrm{RRE} = \arccos\left(\frac{\mathrm{trace}(\textbf{R}^T \cdot \bar{\textbf{R}} - 1)}{2}\right).
\end{equation} 

\emph{Relative Translation Error} (RTE) is the Euclidean distance between estimated and ground-truth translation vectors.
It measures the differences between the predicted and the ground-truth translation vectors.
\begin{equation}
\mathrm{RTE} = \lVert \textbf{t} - \bar{\textbf{t}} \rVert_2.
\end{equation}

\emph{Registration Recall} (RR) on KITTI is defined as the fraction of the point cloud pairs whose RRE and RTE are both below certain thresholds (\ie, $\mathrm{RRE} \tight{<} 5^\circ$ and $\mathrm{RTE} \tight{<} 2\text{m}$).
\begin{equation}
\mathrm{RR} = \frac{1}{M} \sum_{i=1}^{M} \llbracket \mathrm{RRE}_i < 5^{\circ} \land \mathrm{RTE}_i < 2\text{m} \rrbracket.
\end{equation}
Following~\cite{choy2019fully,bai2020d3feat,huang2021predator,lu2021hregnet,yu2021cofinet}, we compute the mean RRE and the mean RTE only for the correctly registered point cloud pairs in KITTI.

\section{Analysis of Cross-Entropy Loss}
\label{supp:celoss}

In this section, we first give an analysis that adopting the cross-entropy loss in multi-label classification problem could suppress the classes with high confidence scores.
Given the input vector $\textbf{y} \in \mathbb{R}^n$ and the label vector $\textbf{g} \in \{0, 1\}^n$, the confidence vector $\textbf{z}$ is computed by adopting a softmax on $\textbf{y}$:
\begin{equation}
z_i = \frac{\exp(y_i)}{\sum_{j=1}^{n} \exp(y_j)}.
\end{equation}
The cross-entropy loss is computed as:
\begin{equation}
\mathcal{L} = -\sum_{i=1}^n g_i\log(z_i).
\end{equation}
And the gradient vector $\textbf{d}$ of $\textbf{y}$ is computed as:
\begin{equation}
d_i = \frac{\partial \mathcal{L}}{\partial y_i} = (\sum_{j=1}^n g_j)z_i - g_i.
\end{equation}
The zero point of $d_i$ \wrt $z_i$ is $c_i = g_i / \sum_{j=1}^n g_j$.
If there are multiple positive classes, we have $0 < c_i < 1$ for each positive class as $\sum_{j=1}^n g_j > 1$.
Hence $y_i$ will be increased if $z_i < c_i$ ($d_i < 0$), and be reduced if $z_i > c_i$ ($d_i > 0$).
This indicates that the cross-entropy loss will suppress the positive classes with higher confidence scores during training.

Now we go back to context of superpoint matching.
To supervise the superpoint matching, CoFiNet~\cite{yu2021cofinet} adopts a cross-entropy loss with an optimal transport layer, which formulates the superpoint matching as a multi-class classification problem for each superpoint.
The ground-truth superpoint correspondences are determined by whether their neighboring point patches overlap.
In practice, one patch usually overlaps with multiple patches in the other point cloud, so superpoint matching is a multi-class classification problem.
According to the analysis above, the positive matches with higher confidence scores will be suppressed by the cross-entropy loss, which hinders the model from extracting reliable superpoint correspondences.
CoFiNet~\cite{yu2021cofinet} further designs a reweighting method which gives better zero points for the gradients, but the problem cannot be solved completely.
On the contrary, our overlap-aware circle loss supervises the superpoint matching in a metric learning manner, which avoids this issue.

\section{Additional Experiments}
\label{supp:additional-exp}

In this section, we conduct more experiments to evaluate our method.
In \cref{supp:detailed-3dmatch}, we provide more detailed comparison on 3DMatch and 3DLoMatch.
In \cref{supp:deep-robust}, we compare our method with recent deep robust estimators.
In \cref{supp:additional-ablation}, we conduct more ablation studies to better understand our design choices.

\subsection{Detailed Results on 3DMatch}
\label{supp:detailed-3dmatch}

\paragraph{Registration results with different overlaps.}

We first compare the performance of the models with vanilla self-attention and our geometric self-attention under different overlap ratios on 3DMatch and 3DLoMatch.
As shown in \cref{table:overlap-level}, our method consistently outperforms the vanilla self-attention counterpart on all the metrics in all levels of overlap ratio.
The gains are greater when the overlap ratio is below $30\%$, demonstrating our method is more robust in low-overlap scenarios.

\vspace{-10pt}
\paragraph{Scene-wise registration results.}

We present the scene-wise registration results on 3DMatch and 3DLoMatch in \cref{table:scene-wise}.
Following~\cite{choy2019fully,bai2020d3feat,huang2021predator}, we report mean median RRE and RTE for the successfully registered point cloud pairs.
For the registration recall, our method outperforms the baselines in most scenes on 3DMatch, especially the hard scenes such as \texttt{Home\_2} and \texttt{Lab}.
And it surpasses the baselines by a large margin in all scenes on 3DLoMatch.
Moreover, our GeoTransformer also achieves consistently superior results on the rotation and translation errors.

\begin{table}[!t]
\scriptsize
\centering
\begin{tabular}{c|ccc|ccc}
\toprule
 \multirow{2}{*}{Overlap} & \multicolumn{3}{c|}{Vanilla Self-attention} & \multicolumn{3}{c}{Geometric Self-attention} \\
  & PIR(\%) & IR(\%) & RR(\%) & PIR(\%) & IR(\%) & RR(\%) \\
\midrule
90\%$-$100\% & 0.974 & 0.829 & 1.000 & 0.989 & 0.894 & 1.000 \\
80\%$-$90\% & 0.948 & 0.787 & 1.000 & 0.969 & 0.859 & 1.000 \\
70\%$-$80\% & 0.902 & 0.731 & 0.931 & 0.935 & 0.815 & 0.931 \\
60\%$-$70\% & 0.884 & 0.686 & 0.933 & 0.939 & 0.783 & 0.946 \\
50\%$-$60\% & 0.843 & 0.644 & 0.957 & 0.913 & 0.750 & 0.970 \\
40\%$-$50\% & 0.787 & 0.579 & 0.935 & 0.867 & 0.689 & 0.944 \\
30\%$-$40\% & 0.716 & 0.523 & 0.917 & 0.818 & 0.644 & 0.940 \\
20\%$-$30\% & 0.560 & 0.406 & 0.781 & 0.666 & 0.518 & 0.839 \\
10\%$-$20\% & 0.377 & 0.274 & 0.639 & 0.466 & 0.372 & 0.705 \\
\bottomrule
\end{tabular}
\caption{
Comparison of the models with the vanilla self-attention and the geometric self-attention under different overlap ratios.
The results are reported on the union of 3DMatch and 3DLoMatch.
}
\label{table:overlap-level}
\end{table}

\begin{table}[!t]
\scriptsize
\centering
\begin{tabular}{l|ccc}
\toprule
Model & RTE(cm) & RRE($^{\circ}$) & RR(\%) \\
\midrule
\multicolumn{4}{c}{3DMatch} \\
\midrule
FCGF+3DRegNet~\cite{pais20203dregnet} & 8.13 & 2.74 & 77.8 \\
FCGF+DGR~\cite{choy2020deep} & 7.36 & 2.33 & 86.5 \\
FCGF+PointDSC~\cite{bai2021pointdsc} & \underline{6.55} & \underline{2.06} & \underline{93.3} \\
FCGF+DHVR~\cite{lee2021deep} & 6.61 & 2.08 & 91.4 \\
PCAM~\cite{cao2021pcam} & $\sim$7 & 2.16 & 92.4 \\
GeoTransformer (\emph{ours}, LGR) & \textbf{5.69} & \textbf{1.98} & \textbf{95.0} \\
\midrule
\multicolumn{4}{c}{3DLoMatch} \\
\midrule
FCGF+PointDSC~\cite{bai2021pointdsc} & \underline{10.50} & \underline{3.82} & \underline{56.2} \\
FCGF+DHVR~\cite{lee2021deep} & 11.76 & 3.88 & 55.6 \\
GeoTransformer (\emph{ours}, LGR) & \textbf{8.55} & \textbf{2.98} & \textbf{77.5} \\
\midrule
\multicolumn{4}{c}{KITTI} \\
\midrule
FCGF+DGR~\cite{choy2020deep} & 21.7 & 0.34 & 96.9 \\
FCGF+PointDSC~\cite{bai2021pointdsc} & 20.9 & 0.33 & 98.2 \\
FCGF+DHVR~\cite{lee2021deep} & 19.8 & \underline{0.29} & \underline{99.1} \\
PCAM~\cite{cao2021pcam} & \underline{$\sim$8} & 0.33 & 97.2 \\
GeoTransformer (\emph{ours}, LGR) & \textbf{6.5} & \textbf{0.24} & \textbf{99.5} \\
\bottomrule
\end{tabular}
\caption{
Comparison with deep robust estimators on 3DMatch and KITTI.
The RTE of PCAM is rounded to centimeter in the original paper~\cite{cao2021pcam}.
}
\label{table:supp-direct}
\end{table}

\begin{table*}[!t]
\scriptsize
\setlength{\tabcolsep}{2.5pt}
\centering
\begin{tabular}{l|ccccccccc|ccccccccc}
\toprule
\multirow{2}{*}{Model} & \multicolumn{9}{c|}{3DMatch} & \multicolumn{9}{c}{3DLoMatch} \\
 & Kitchen & Home\_1 & Home\_2 & Hotel\_1 & Hotel\_2 & Hotel\_3 & Study & Lab & Mean & Kitchen & Home\_1 & Home\_2 & Hotel\_1 & Hotel\_2 & Hotel\_3 & Study & Lab & Mean \\
\midrule
\multicolumn{19}{c}{\emph{Registration Recall} (\%) $\uparrow$} \\
\midrule
3DSN~\cite{gojcic2019perfect} & 90.6 & 90.6 & 65.4 & 89.6 & 82.1 & 80.8 & 68.4 & 60.0 & 78.4 & 51.4 & 25.9 & 44.1 & 41.1 & 30.7 & 36.6 & 14.0 & 20.3 & 33.0 \\
FCGF~\cite{choy2019fully} & \underline{98.0} & 94.3 & 68.6 & 96.7 & \underline{91.0} & \underline{84.6} & 76.1 & 71.1 & 85.1 & 60.8 & 42.2 & 53.6 & 53.1 & 38.0 & 26.8 & 16.1 & 30.4 & 40.1 \\
D3Feat~\cite{bai2020d3feat} & 96.0 & 86.8 & 67.3 & 90.7 & 88.5 & 80.8 & 78.2 & 64.4 & 81.6 & 49.7 & 37.2 & 47.3 & 47.8 & 36.5 & 31.7 & 15.7 & 31.9 & 37.2 \\
Predator~\cite{huang2021predator} & 97.6 & \underline{97.2} & \underline{74.8} & \textbf{98.9} & \textbf{96.2} & \textbf{88.5} & 85.9 & 73.3 & 89.0 & 71.5 & 58.2 & 60.8 & 77.5 & 64.2 & 61.0 & 45.8 & 39.1 & 59.8 \\
CoFiNet~\cite{yu2021cofinet} & 96.4 & \textbf{99.1} & 73.6 & \underline{95.6} & \underline{91.0} & \underline{84.6} & \textbf{89.7} & \underline{84.4} & \underline{89.3} & \underline{76.7} & \underline{66.7} & \underline{64.0} & \underline{81.3} & \underline{65.0} & \underline{63.4} & \underline{53.4} & \underline{69.6} & \underline{67.5} \\
P2PNet (\emph{ours}) & \textbf{98.9} & \underline{97.2} & \textbf{81.1} & \textbf{98.9} & 89.7 & \textbf{88.5} & \underline{88.9} & \textbf{88.9} & \textbf{91.5} & \textbf{85.9} & \textbf{73.5} & \textbf{72.5} & \textbf{89.5} & \textbf{73.2} & \textbf{66.7} & \textbf{55.3} & \textbf{75.7} & \textbf{74.0} \\
\midrule
\multicolumn{19}{c}{\emph{Relative Rotation Error} ($^{\circ}$) $\downarrow$} \\
\midrule
3DSN~\cite{gojcic2019perfect} & 1.926 & 1.843 & 2.324 & 2.041 & 1.952 & 2.908 & 2.296 & 2.301 & 2.199 & 3.020 & 3.898 & 3.427 & 3.196 & 3.217 & 3.328 & 4.325 & 3.814 & 3.528 \\
FCGF~\cite{choy2019fully} & \textbf{1.767} & 1.849 & \underline{2.210} & 1.867 & 1.667 & 2.417 & \underline{2.024} & \underline{1.792} & \underline{1.949} & \underline{2.904} & 3.229 & 3.277 & 2.768 & \underline{2.801} & \underline{2.822} & 3.372 & 4.006 & 3.147 \\
D3Feat~\cite{bai2020d3feat} & 2.016 & 2.029 & 2.425 & 1.990 & 1.967 & 2.400 & 2.346 & 2.115 & 2.161 & 3.226 & 3.492 & 3.373 & 3.330 & 3.165 & 2.972 & 3.708 & 3.619 & 3.361 \\
Predator~\cite{huang2021predator} & 1.861 & \underline{1.806} & 2.473 & 2.045 & \underline{1.600} & 2.458 & 2.067 & 1.926 & 2.029 & 3.079 & \underline{2.637} & \underline{3.220} & \underline{2.694} & 2.907 & 3.390 & \underline{3.046} & 3.412 & \underline{3.048} \\
CoFiNet~\cite{yu2021cofinet} & 1.910 & 1.835 & 2.316 & \underline{1.767} & 1.753 & \underline{1.639} & 2.527 & 2.345 & 2.011 & 3.213 & 3.119 & 3.711 & 2.842 & 2.897 & 3.194 & 4.126 & \underline{3.138} & 3.280 \\
P2PNet (\emph{ours}) & \underline{1.797} & \textbf{1.353} & \textbf{1.797} & \textbf{1.528} & \textbf{1.328} & \textbf{1.571} & \textbf{1.952} & \textbf{1.678} & \textbf{1.625} & \textbf{2.356} & \textbf{2.305} & \textbf{2.541} & \textbf{2.455} & \textbf{2.490} & \textbf{2.504} & \textbf{3.010} & \textbf{2.716} & \textbf{2.547} \\
\midrule
\multicolumn{19}{c}{\emph{Relative Translation Error} (m) $\downarrow$} \\
\midrule
3DSN~\cite{gojcic2019perfect} & 0.059 & 0.070 & 0.079 & 0.065 & 0.074 & 0.062 & 0.093 & 0.065 & 0.071 & 0.082 & 0.098 & 0.096 & 0.101 & \underline{0.080} & 0.089 & 0.158 & \underline{0.120} & 0.103 \\
FCGF~\cite{choy2019fully} & 0.053 & 0.056 & 0.071 & \underline{0.062} & 0.061 & 0.055 & 0.082 & 0.090 & 0.066 & 0.084 & 0.097 & \underline{0.076} & 0.101 & 0.084 & 0.077 & 0.144 & 0.140 & 0.100 \\
D3Feat~\cite{bai2020d3feat} & 0.055 & 0.065 & 0.080 & 0.064 & 0.078 & \underline{0.049} & 0.083 & 0.064 & 0.067 & 0.088 & 0.101 & 0.086 & \underline{0.099} & 0.092 & \underline{0.075} & 0.146 & 0.135 & 0.103 \\
Predator~\cite{huang2021predator} & 0.048 & \underline{0.055} & 0.070 & 0.073 & 0.060 & 0.065 & \underline{0.080} & \underline{0.063} & 0.064 & 0.081 & 0.080 & 0.084 & \underline{0.099} & 0.096 & 0.077 & \textbf{0.101} & 0.130 & \underline{0.093} \\
CoFiNet~\cite{yu2021cofinet} & \underline{0.047} & 0.059 & \underline{0.063} & 0.063 & \underline{0.058} & \textbf{0.044} & 0.087 & 0.075 & \underline{0.062} & \underline{0.080} & \underline{0.078} & 0.078 & \underline{0.099} & 0.086 & 0.077 & 0.131 & 0.123 & 0.094 \\
P2PNet (\emph{ours}) & \textbf{0.042} & \textbf{0.046} & \textbf{0.059} & \textbf{0.055} & \textbf{0.046} & 0.050 & \textbf{0.073} & \textbf{0.053} & \textbf{0.053} & \textbf{0.062} & \textbf{0.070} & \textbf{0.071} & \textbf{0.080} & \textbf{0.075} & \textbf{0.049} & \underline{0.107} & \textbf{0.083} & \textbf{0.074} \\
\bottomrule
\end{tabular}
\caption{
Scene-wise registration results on 3DMatch and 3DLoMatch.
}
\label{table:scene-wise}
\end{table*}

\subsection{Comparison with Deep Robust Estimators}
\label{supp:deep-robust}

We further compare with recent deep robust estimators: 3DRegNet~\cite{pais20203dregnet}, DGR~\cite{choy2020deep}, PointDSC~\cite{bai2021pointdsc}, DHVR~\cite{lee2021deep} and PCAM~\cite{cao2021pcam} on 3DMatch and KITTI.
Following common practice, we report RTE, RRE and RR on both benchmarks.
Here RR is defined as in \cref{supp:metrics-kitti}.
The RTE threshold is $30\text{cm}$ on 3DMatch and $60\text{cm}$ on KITTI, while the RRE threshold is $15^{\circ}$ on 3DMatch and $5^{\circ}$ on KITTI.
As shown in \cref{table:supp-direct}, our method outperforms all the baselines on both benchmarks.
Although different correspondence extractors are used, these results can already demonstrate the superiority of GeoTransformer.
It is noteworthy that our LGR is parameter-free and does not require training a specific network, which contributes to faster registration speed (0.013s \vs 0.08s~\cite{bai2021pointdsc} in our experiments).

\subsection{Additional Ablation Studies}
\label{supp:additional-ablation}



\begin{table}[!t]
\scriptsize
\centering
\begin{tabular}{l|cc|cc}
\toprule
 \multirow{2}{*}{Model} & \multicolumn{2}{c|}{3DMatch} & \multicolumn{2}{c}{3DLoMatch} \\
  & original & rotated & original & rotated \\
\midrule
(a) self-attention w/ ACE & 89.3 & 87.2{\tiny \ -2.1} & 69.3 & 67.4{\tiny \ -1.9} \\
(b) self-attention w/ RCE & 88.5 & 88.5{\tiny \ same} & 68.7 & 68.7{\tiny \ same} \\
(c) geometric self-attention & 91.5 & 91.4{\tiny \ -0.1} & 74.0 & 73.8{\tiny \ -0.2} \\
\bottomrule
\end{tabular}
\caption{
Ablation experiments with rotated superpoints.
}
\label{table:ablation-study-rotated}
\end{table}

\paragraph{Transformation invariance.}

We first evaluate the transformation invariance of different positional embeddings in \cref{table:ablation-study-rotated}.
For each model, we randomly apply arbitrary rotations to the \emph{superpoints} when computing the superpoint embeddings.
Among all the variants, enlarged rotations severely degrade the performance of (a) self-attention with absolute coordinate embedding, which indicates the lack of transformation variance in it.
Surprisingly, the performance of (b) self-attention with relative coordinate embedding is quite stable.
However, after masking the relative coorinate embedding out, we find that the results of this model still remain the same, which means the relative coordinate embedding contributes little to the final performance during testing.
In constrast, our (c) geometric self-attention shows strong invariance to rigid transformation.



\begin{table}[!t]
\scriptsize
\setlength{\tabcolsep}{3pt}
\centering
\begin{tabular}{l|cccc|cccc}
\toprule
 \multirow{2}{*}{Model} & \multicolumn{4}{c|}{3DMatch} & \multicolumn{4}{c}{3DLoMatch} \\
  & PIR & FMR & IR & RR & PIR & FMR & IR & RR \\
\midrule
(a) distance only & 84.9 & 98.0 & 69.1 & 90.7 & 50.6 & 85.8 & 40.3 & 72.1 \\
(b) $k=1$ angles & 86.5 & 97.9 & 70.6 & 91.0 & 54.6 & 87.1 & 42.7 & 73.1 \\
(c) $k=2$ angles & 86.1 & 97.9 & 70.4 & 91.3 & 55.0 & 88.2 & 43.5 & 73.5 \\
(d) $k=3$ angles & 86.1 & 97.7 & 70.3 & 91.5 & 54.9 & 88.1 & 43.3 & 74.0 \\
(e) $k=4$ angles & 86.6 & 98.0 & 70.7 & 91.7 & 55.1 & 88.4 & 43.5 & 74.2 \\
\midrule
(f) max pooling & 86.1 & 97.7 & 70.3 & 91.5 & 54.9 & 88.1 & 43.3 & 74.0 \\
(g) average pooling & 86.3 & 98.0 & 70.2 & 91.3 & 54.6 & 87.3 & 42.8 & 74.0 \\
\midrule
(h) w/ dual-normalization & 86.1 & 97.7 & 70.3 & 91.5 & 54.9 & 88.1 & 43.3 & 74.0 \\
(i) w/o dual-normalization & 86.2 & 97.7 & 70.3 & 91.0 & 53.5 & 87.9 & 42.8 & 73.8 \\
\bottomrule
\end{tabular}
\caption{Additional ablation experiments.}
\label{table:ablation-study-supp}
\end{table}


\vspace{-10pt}
\paragraph{Geometric structure embedding.}

Next, we study the design of geometric structure embedding.
We first vary the number of nearest neighbors for computing the triplet-wise angular embedding.
As shown in~\cref{table:ablation-study-supp}(top), increasing the neighbors slightly improves the registration recall, but also requires more computation. To better balance accuracy and speed, we select $k \tight{=} 3$ in our experiments unless otherwise noted.
We then replace max pooling with average pooling when aggregating the triplet-wise angular embedding in Eq.~(5) of our main paper.
From~\cref{table:ablation-study-supp}(middle), the results of two pooling methods are very close and max pooling performs slightly better than average pooling.

\vspace{-10pt}
\paragraph{Dual-normalization.}

We then investigate the effectiveness of the dual-normalization operation in the superpoint matching module.
As shown in \cref{table:ablation-study-supp}(bottom), it slightly improves the accuracy of the superpoint correspondences in low-overlap scenarios.
As there is less overlapping context when the overlapping area is small, it is much easier to extract outlier matches between the less geometrically discriminative patches.
The dual-normalization operation can mitigate this issue and slightly improves the performance.

\vspace{-10pt}
\paragraph{Pose refinement.}

At last, we evaluate the impact of the pose refinement in LGR.
As shown in \cref{fig:refinement}, the registration recall consistently improves with more iterations and gets saturated after about $5$ iterations.
To better balance accuracy and speed, we choose $5$ iterations in the experiments.

\input{figures/refinement}

\section{Limitations}

GeoTransformer relies on uniformly downsampled superpoints to hierarchically extract correspondences. However, there could be numerous superpoints if the input point clouds cover a large area, which will cause huge memory usage and computational cost. In this case, we might need to carefully select the downsampling rate to balance performance and efficiency.

Besides, it is inflexible to uniformly sample superpoints (patches). In practice, it is common that a single object is decomposed into multiple patches, which could be easily registered as a whole. So we think that it is a very promising topic to integrate point cloud registration with semantic scene understanding tasks (\eg, semantic segmentation and object detection), which converts scene registration into object registration. We will leave this for future work.

\section{Qualitative Results}

We provide more qualitative results on 3DLoMatch in \cref{fig:supp-gallery-3dmatch}. The registration results from Predator~\cite{huang2021predator} and CoFiNet~\cite{yu2021cofinet} are also shown for comparison. Here Predator and CoFiNet use RANSAC-$50k$ for registration, while LGR is used in GeoTransformer. Our method performs quite well in these low-overlap cases. It is noteworthy that our method can distinguish similar objects at different positions (see the comparison of CoFiNet and GeoTransformer in the $4^{\text{th}}$ and $6^{\text{th}}$ rows) thanks to the transformation invariance obtained from the geometric self-attention. \cref{fig:supp-gallery-kitti} visualizes the registration results of GeoTransformer in the bird's-eye view on KITTI. It can be observed that our method attains very accurate registration even without using RANSAC.


We also provide some failure cases of our method on 3DLoMatch in \cref{fig:supp-gallery-failure}. Generally, if the overlaping region between two point clouds is small and geometrically indiscriminative (\eg, wall, ceiling and floor) or the non-overlapping region is relatively complicated, the registration could fail. A commonality of these cases is that they cannot provide adequate geometric cues to detect overlapping area and extract reliable correspondences. A possible solution could combine the information from multiple point clouds. We will leave this for future work.

\input{figures/supp-gallery-3dmatch}

\input{figures/supp-gallery-kitti}

\input{figures/supp-gallery-failure}

\end{appendix}

%% file: figures/supp-cross-att.tex

\begin{figure}[t]
  \centering
  \begin{overpic}[width=1.0\linewidth]{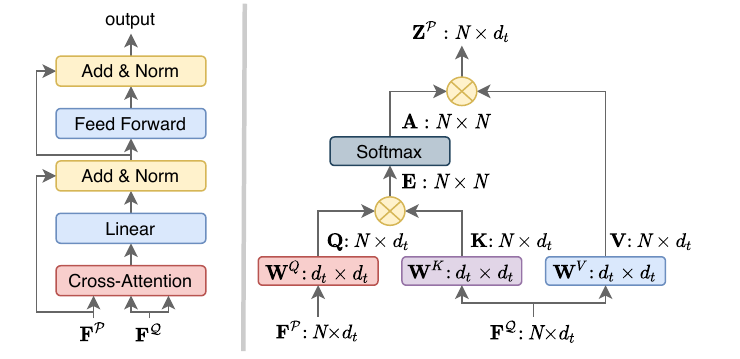}
  \end{overpic}
  \vspace{-20pt}
  \caption{Left: The structure of feature-based cross-attention module. Right: The computation graph of cross-attention.}
  \label{fig:supp-cross-att}
  \vspace{-10pt}
\end{figure}

%% file: figures/refinement.tex

\begin{figure}[t]
  \centering
  \begin{overpic}[width=0.8\linewidth]{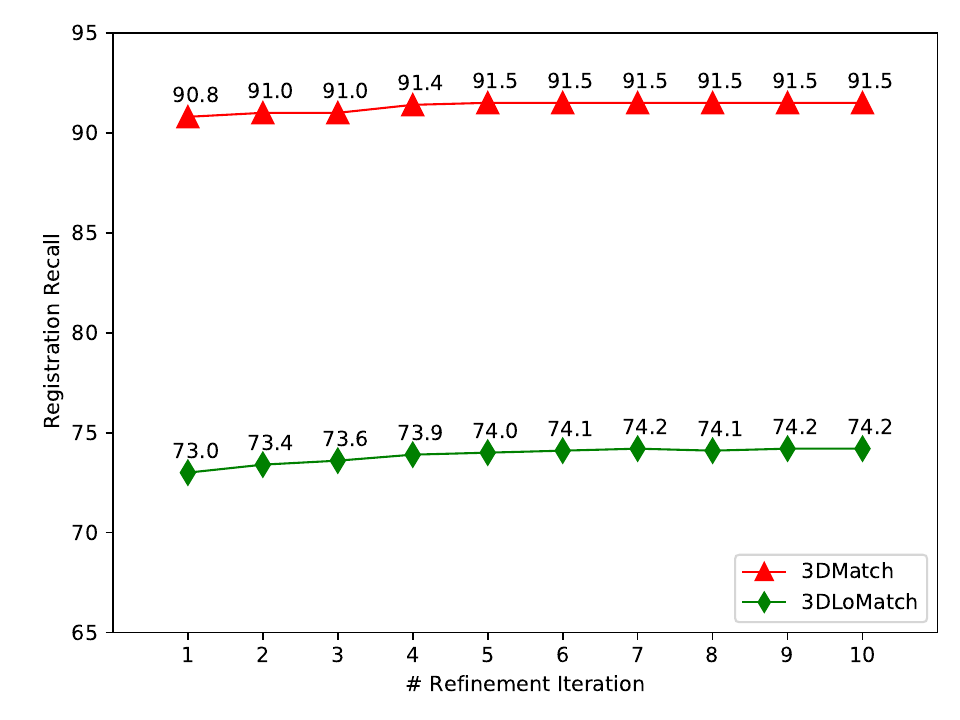}
  \end{overpic}
  \vspace{-15pt}
  \caption{Ablation experiments on the pose refinement.}
  \label{fig:refinement}
  \vspace{-15pt}
\end{figure}

%% file: figures/supp-gallery-3dmatch.tex

\begin{figure*}[t]
  \begin{overpic}[width=1.0\linewidth]{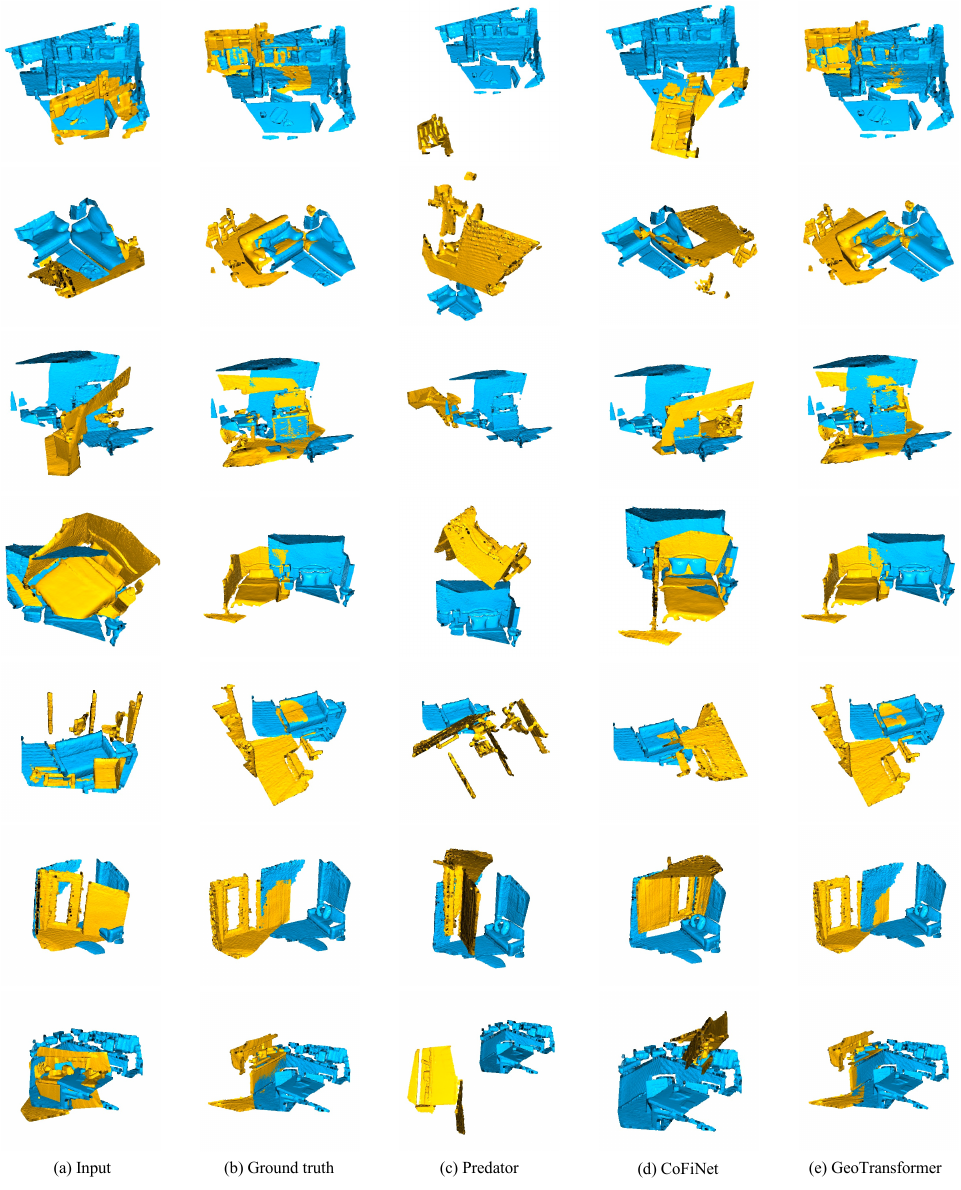}
  \end{overpic}
  \caption{Registration results on 3DMatch and 3DLoMatch.
  }
  \label{fig:supp-gallery-3dmatch}
\end{figure*}

%% file: figures/supp-gallery-kitti.tex

\begin{figure*}[t]
  \begin{overpic}[width=1.0\linewidth]{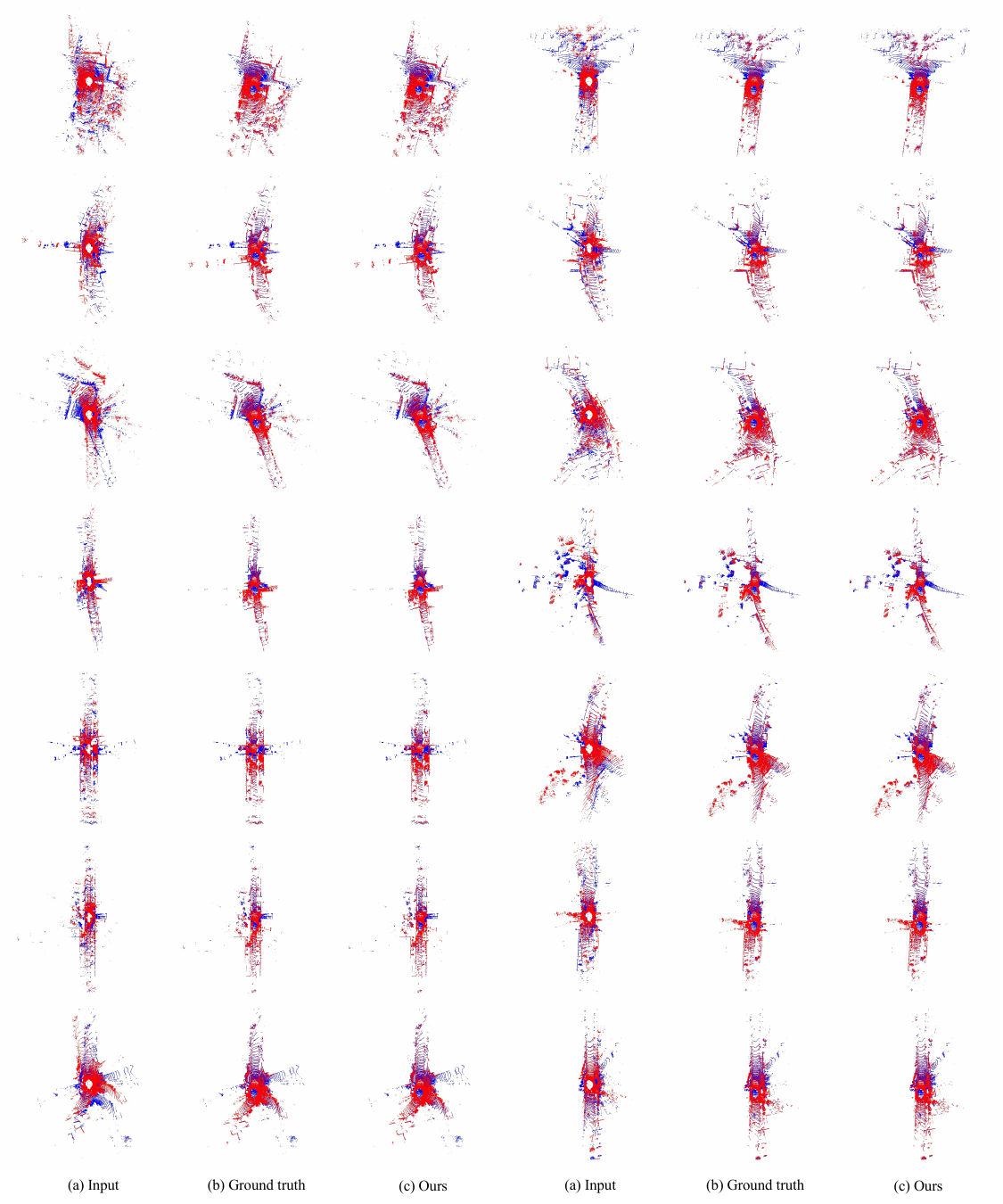}
  \end{overpic}
  \caption{Registration results on KITTI odometry.
  }
  \label{fig:supp-gallery-kitti}
\end{figure*}

%% file: figures/supp-gallery-failure.tex

\begin{figure*}[t]
  \begin{overpic}[width=1.0\linewidth]{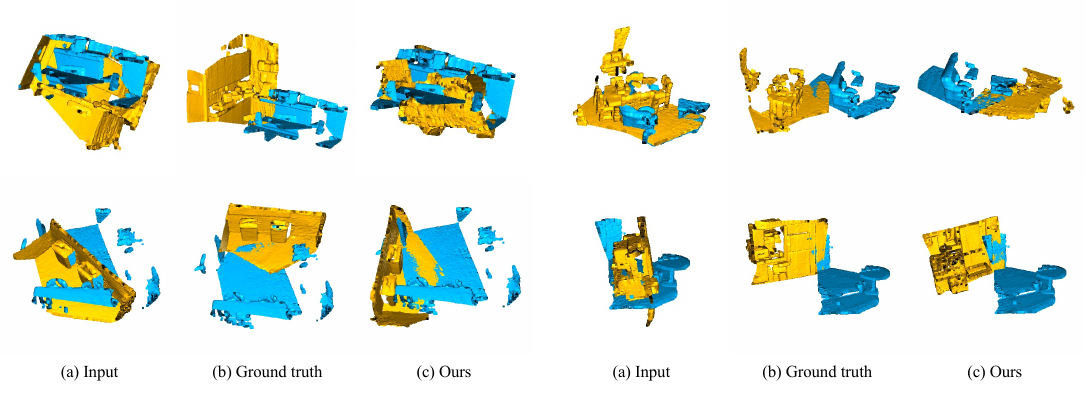}
  \end{overpic}
  \vspace{-10pt}
  \caption{Failed cases on 3DLoMatch.
  }
  \label{fig:supp-gallery-failure}
  \vspace{-10pt}
\end{figure*}